\documentclass[letterpaper]{article} % DO NOT CHANGE THIS
\usepackage{aaai25}  % DO NOT CHANGE THIS
\usepackage{times}  % DO NOT CHANGE THIS
\usepackage{helvet}  % DO NOT CHANGE THIS
\usepackage{courier}  % DO NOT CHANGE THIS
\usepackage[hyphens]{url}  % DO NOT CHANGE THIS
\usepackage{graphicx} % DO NOT CHANGE THIS
\usepackage{natbib}  % DO NOT CHANGE THIS AND DO NOT ADD ANY OPTIONS TO IT
\usepackage{caption} % DO NOT CHANGE THIS AND DO NOT ADD ANY OPTIONS TO IT
\usepackage{algorithm}
\usepackage{algorithmic}
\usepackage{latexsym}
\usepackage[utf8]{inputenc}
\usepackage{microtype}
\usepackage{inconsolata}
\usepackage{amsmath}
\usepackage{amssymb}
\usepackage{xspace}
\usepackage{booktabs}
\usepackage{multirow}
\usepackage{graphicx}
\usepackage{tabularx}
\usepackage{subcaption}
\usepackage{tcolorbox}

\newcommand{\paratitle}[1]{\vspace{1.5ex}\noindent\textbf{#1}}
\newcommand{\ie}{\emph{i.e.,}\xspace}

\newcommand{\eg}{\emph{e.g.,}\xspace}

\newcommand{\ignore}[1]{}

\newcommand{\OURS}{GPO\xspace}

\usepackage{newfloat}
\usepackage{listings}
\DeclareCaptionStyle{ruled}{labelfont=normalfont,labelsep=colon,strut=off} % DO NOT CHANGE THIS
\floatstyle{ruled}
\newfloat{listing}{tb}{lst}{}
\floatname{listing}{Listing}
\title{Unleashing the Potential of Large Language Models as Prompt Optimizers: Analogical Analysis with Gradient-based Model Optimizers}
\author{
    Xinyu Tang\textsuperscript{\rm{1}\thanks{\ \ Equal contribution.}},
    Xiaolei Wang\textsuperscript{\rm{1}\footnotemark[1]},
    Wayne Xin Zhao\textsuperscript{\rm{1}\thanks{\ \ Corresponding author.}},
    Siyuan Lu\textsuperscript{\rm{2}\ },
    Yaliang Li\textsuperscript{\rm{3}\ },
    Ji-Rong Wen\textsuperscript{\rm{1}}
}
\affiliations{
    \textsuperscript{1}Gaoling School of Artificial Intelligence, Renmin University of China \\
    \textsuperscript{2}International School, Beijing University of Posts and Telecommunications \\
    \textsuperscript{3}Alibaba Group \\
    {txy20010310@163.com, wxl1999@foxmail.com, batmanfly@gmail.com}\\
}

\usepackage{bibentry}
\begin{document}

\maketitle

\begin{abstract}
Automatic prompt optimization is an important approach to improving the performance of large language models~(LLMs).
Recent research demonstrates the potential of using LLMs as prompt optimizers, which can generate improved task prompts via iterative refinement. 
In this paper, we propose a novel perspective to investigate the design of LLM-based prompt optimizers, by drawing an analogy with gradient-based model optimizers. 
To connect these two approaches, we identify two pivotal factors in model parameter learning: \emph{update direction} and \emph{update method}. 
By systematically analyzing a rich set of improvement strategies on the two aspects, we further develop a capable \textbf{G}radient-inspired LLM-based \textbf{P}rompt \textbf{O}ptimizer called \textbf{\OURS}.
At each step, it first retrieves relevant prompts from the optimization trajectory as the update direction.
Then, it utilizes the generation-based refinement strategy to perform the update, while controlling the edit distance through a cosine-based decay strategy.
Extensive experiments demonstrate the effectiveness and efficiency of {\OURS}.
In particular, \OURS brings an additional improvement of up to 56.8\% on Big-Bench Hard and 62.6\% on MMLU compared to baseline methods.
The code is available at \url{https://github.com/RUCAIBox/GPO}.
\end{abstract}

% \begin{links}
% \link{Code}{https://github.com/RUCAIBox/GPO}
% \end{links}

\section{Introduction}

Nowadays, prompting has become the pivotal approach to unleashing the power of large language models~(LLMs)~\cite{survey-llm}.
However, prompt engineering is not easy and requires extensive trial-and-error efforts since LLMs are sensitive to prompts~\cite{Fantastically-Lu-2022-ACL, REAR-Wang-EMNLP, DAWN-ICL}.
Although general guidelines for high-quality prompts exist~\cite{Large-Kojima-2022-NeurIPS, Prompt-Amatriain-2024-arXiv}, they cannot always lead to optimal task performance.

To improve the task performance of LLMs, \emph{automatic prompt optimization} has been proposed~\cite{Large-Zhou-2023-ICLR}.
Early work either performs discrete optimization through methods like reinforcement learning~\cite{TEMPERA-Zhang-2023-ICLR} or performs continuous optimization in the embedding space~\cite{InstructZero-Chen-2023-arXiv, UniCRS-KDD, CFCRS-KDD}.
However, they often require access to the logits or internal states of LLMs, which is infeasible for those only accessible through APIs.
In addition, they need to be specially trained for each task.
Considering these issues, recent work proposes to model the optimization problem in natural language and using LLMs as \emph{prompt optimizers} due to their strong understanding~\cite{Over-Zhan-NIPS} and generation capabilities~\cite{Prompt-Tobias-2024-arXiv}.
In this approach, LLMs perform optimization with only outputs from the task model and quickly adapt to various tasks without training.
However, such a method raises a new challenge for the design of \textit{meta-prompt}, which is the prompt for LLMs to perform prompt optimization. 

\begin{figure}[t!]
    \centering
    \includegraphics[width=\linewidth]{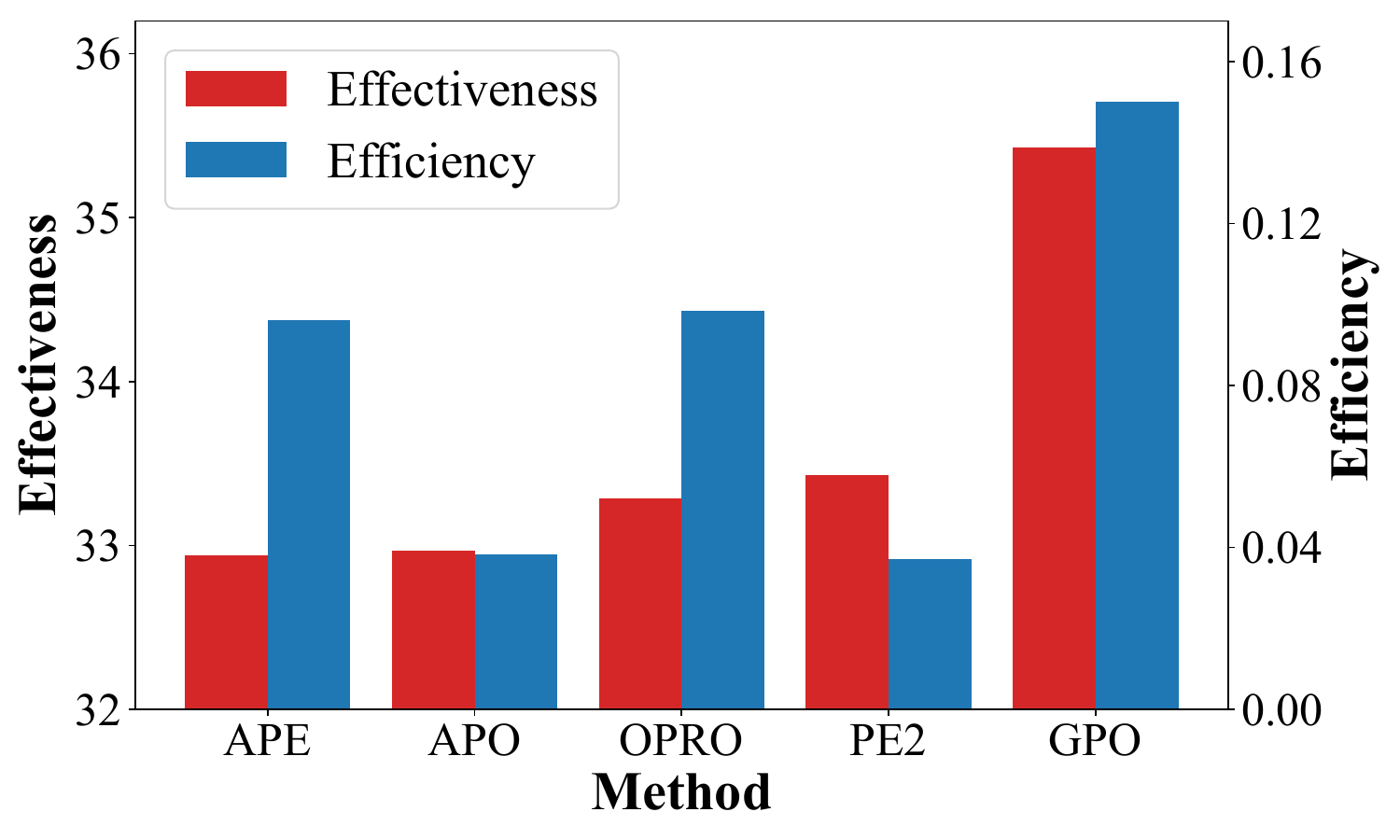}
    \caption{Comparisons of \OURS to existing LLM-based prompt optimizers in terms of effectiveness~(Accuracy) and efficiency~(improvement per dollar spent on API) on BBH.
    } 
    \label{fig:2e}
\end{figure}

\begin{figure*}[ht!]
    \centering
    \includegraphics[width=0.95 \linewidth]{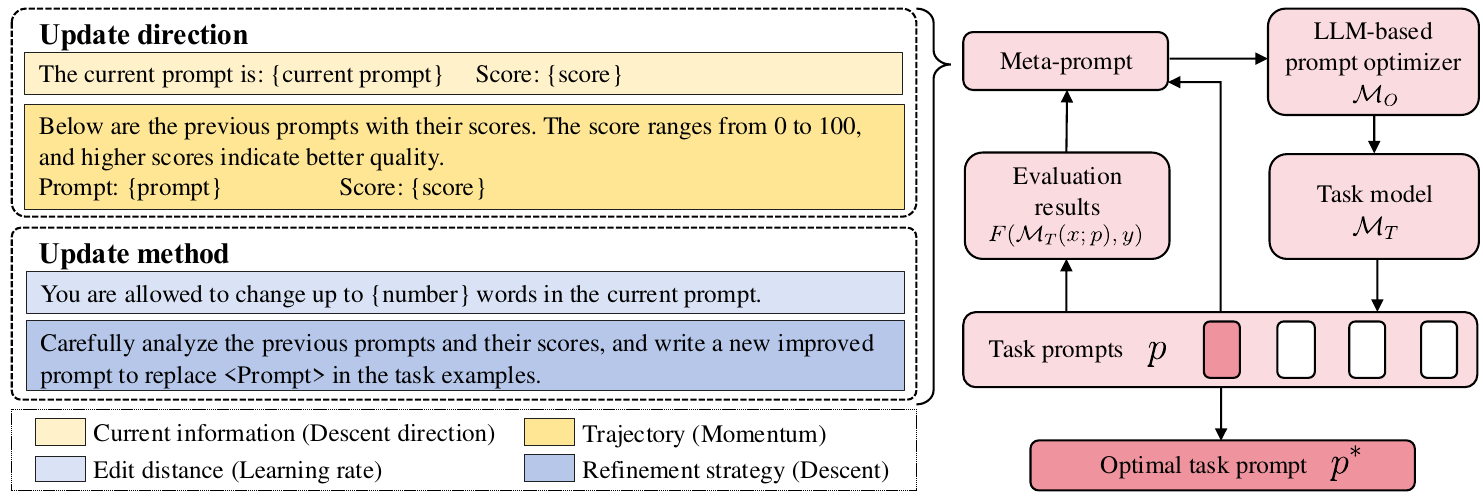}
    \caption{The overview of the \OURS framework. {``Current information'', ``Trajectory'', ``Edit distance'', and ``Refinement strategy'' are concepts of LLM-based prompt optimizers, which can correspond to ``Descent direction'', ``Momentum'', ``Learning rate'', and ``Descent'' in gradient-based model optimizers.}}
    \label{fig:GPO}
\end{figure*}

To tackle this issue, we aim to investigate the design of meta-prompts.
Existing methods for creating meta-prompts typically involve either manual human effort~\cite{Large-Yang-2023-arXiv} or heuristic algorithms~\cite{Promptbreeder-Fernando-2023-arXiv}.
Despite the flexibility, these studies still lack principled guidelines about their designs.
Our work is inspired by the great success of gradient-based optimizers in model optimization, which have been systemically studied in both theory and practice~\cite{A-Sun-2020-IEEE}.
Since both optimizers aim at enhancing model performance through iterative optimization, it is feasible to connect the two different approaches via analogical analysis.
In this way, we can borrow the theoretical framework and extensive research of gradient-based model optimizers to enhance LLM-based prompt optimizers.

Therefore, in this paper, we propose a comprehensive analogy framework for the two key factors (\ie \textit{update direction} and \textit{update method}) in LLM-based optimizers.
As illustrated in Table~\ref{tab:glossary}, for update direction, we analogize descent direction to the information of the current prompt and momentum to the information of the historical prompt, while for the update method, we analogize the learning rate to edit distance and gradient descent to the refinement strategy.
{Based on such a framework, we conduct systematic empirical studies on various implementations for each component and report their experimental results with detailed analysis.}

Based on the findings from our systematic analysis, we further develop a capable \textbf{G}radient-inspired LLM-based \textbf{P}rompt \textbf{O}ptimizer called {\textbf{\OURS}}, with the best implementation for each component.
Figure~\ref{fig:GPO} illustrates the overall framework of \OURS.
We evaluate its effectiveness across various LLMs, tasks, and evaluation settings.
When using \texttt{Llama-2-7b-chat} as the task model, the prompts produced by \OURS surpass the instruction ``Let's think step by step'' by 18.5\% on Big-Bench Hard~(BBH) and 7.6\% on MMLU.
{Furthermore, \OURS produces an additional improvement of up to 56.8\% on BBH and 62.6\% on MMLU compared with baseline methods while using fewer tokens.}

Our contributions can be summarized as follows:

$\bullet$ To the best of our knowledge, {this is the first time that a systematic analogy has been conducted for LLM-based prompt optimizers with gradient-based model optimizers.}

$\bullet$ We conduct a comprehensive experimental analysis on the two key factors (\ie update direction and update method) and report several key findings.

$\bullet$ Based on the findings of the systematic analysis, we develop a more effective and efficient LLM-based prompt optimizer, \OURS, which surpasses competitive baseline methods across various LLMs, tasks, and evaluation settings while incurring lower costs.

\section{Related Work}

\subsubsection{Prompt Engineering and Optimization.}
Prompt engineering aims to find suitable prompts for LLMs to perform various tasks. 
To reduce human efforts, researchers have explored automatic prompt optimization, which can be categorized into continuous and discrete optimization methods.
Discrete methods directly optimize the natural language prompts through methods like reinforcement learning~\cite{RLPrompt-Deng-2022-EMNLP, TEMPERA-Zhang-2023-ICLR} and editing~\cite{GPS-Xu-2022-EMNLP, GrIPS-Prasad-2023-EACL}.
In contrast, continuous methods perform optimization in the embedding space of LLMs, allowing for optimization through gradient~\cite{Prefix-Li-2021-ACL}.
We focus on discrete methods, especially LLM-based prompt optimizers.

\subsubsection{LLM-based Prompt Optimizers.}
Due to the unprecedented capabilities of LLMs, recent work starts to utilize them as prompt optimizers.
One line of work~\cite{Connecting-Guo-2023-arxiv, InstOptima-Yang-2023-EMNLP} combines LLMs with evolutionary algorithms to perform prompt optimization.
Another line of work~\cite{Large-Yang-2023-arXiv} aims to adapt concepts and techniques from gradient-based model optimizers (\eg gradient~\cite{Automatic-Pryzant-2023-EMNLP} and momentum~\cite{Large-Yang-2023-arXiv}) to LLM-based prompt optimizers.
However, no comprehensive guidelines exist for using LLMs as prompt optimizers.
We aim to tackle this with a systematic investigation, which is conducted by analogy with gradient-based model optimizers.

\section{Analogical Analysis}
\label{sec:analogical-analysis}
In this section, we present an analogical analysis between model optimization and prompt optimization to build connections and improve existing LLM-based prompt optimizers.

\begin{table}[t]
\centering
\tiny
\resizebox{0.95 \columnwidth}{!}{%
\begin{tabular}{c|cc}
\toprule
\multicolumn{1}{c|}{\textbf{Factor}}                 & \begin{tabular}[c]{@{}c@{}}\textbf{Gradient-based}\\\textbf{model optimizer}\end{tabular} & \begin{tabular}[c]{@{}c@{}}\textbf{LLM-based}\\\textbf{prompt optimizer}\end{tabular} \\
\midrule
\multirow{2}{*}{\begin{tabular}[c]{@{}c@{}}Update\\direction\end{tabular}}        & {Descent direction}        & Current information       \\
                                                                                    & Momentum        & Trajectory       \\
\midrule
\multirow{2}{*}{\begin{tabular}[c]{@{}c@{}}Update\\method\end{tabular}}                                         & Learning rate   & Edit distance \\
                                                                          & Descent         & Refinement strategy \\
\bottomrule
\end{tabular}%
}
\caption{Analogy between glossaries in model optimizer and prompt optimizer.}
\label{tab:glossary}
\end{table}

\subsection{Background}
In this part, we first introduce the task definition of LLM-based prompt optimization, and then establish connections with gradient-based optimizers.

\subsubsection{Task Formulation.}
Prompt optimization aims to find the optimal \textit{task prompt} $p^*$ in the format of natural language that maximizes the performance on a specific task dataset $\mathcal{D}$ when using an LLM as the task model $\mathcal{M}_T$.
To perform such optimization, our idea is to develop a prompt optimizer, which can be built upon some search algorithm (\eg evolutionary algorithms~\cite{Connecting-Guo-2023-arxiv}) or an LLM~\cite{Large-Yang-2023-arXiv}.
In this paper, we focus on using an LLM as the prompt optimizer $\mathcal{M}_O$.
Formally, the problem of prompt optimization can be formulated as:  
\begin{equation}
p^* = \mathop{\arg\max} \limits_{p \sim \mathcal{M}_O} \ \mathbb{E}_{\langle x,y \rangle \in \mathcal{D}} \ [F(\mathcal{M}_T(x;p), y)],
\label{eq:prompt-optimization}
\end{equation}
where $p$ is the prompt generated by the prompt optimizer $\mathcal{M}_O$, $\mathcal{M}_T(x; p)$ represents the output from the task model for input $x$ conditioned on the prompt $p$, and the function $F(\cdot)$ calculates the task performance based on some measurement.

\subsubsection{Revisiting Gradient-based Optimizers.}
Similar to LLM-based prompt optimizer, gradient-based model optimizer aims to find the optimal values of model parameters that minimize the loss function.
In the basic form of gradient descent~\cite{Convex-Boyd-2014-Cambridge}, a single optimization step can be formulated as follows:
\begin{equation}
    \Theta_{k+1} = \Theta_k - \tau_k g_k,
    \label{eq:update}
\end{equation}
where $\Theta_k$ and $\Theta_{k+1}$ are the values of model parameters at the last and current steps, $\tau_k$ and $g_k$ are the learning rate and gradient at the current step.
Gradient descent can be improved by focusing on two elements in the formula: $\tau_k$ and $g_k$.
For $\tau_k$, learning rate schedulers~\cite{A-Gotmare-2019-ICLR} are proposed to dynamically adjust the learning rate.
For $g_k$, the concept of momentum~\cite{On-Sutskever-2013-ICML} is introduced to include historical gradients, and its computation can be expressed as follows: $v_{k+1} = \beta v_k + g_k = \sum_{i=1}^k \beta^{k-i} g_i$, where $\beta$ represents the momentum coefficient.

Despite various gradient-based optimizers, they mainly model two key factors, namely \emph{update direction} (\eg gradient $g_k$) and \emph{update method}  (\eg subtract $\tau_k g_k$).
Our approach is inspired by the observation that existing LLM-based prompt optimization methods also implicitly employ the two aspects (see Table~\ref{tab:glossary}).
However, existing work only initially explores the design of the two key factors, we aim to conduct more in-depth and systematic investigations from these two novel perspectives in the following subsection.

\subsection{Update Direction}
\label{subsec:source}
The update direction refers to the adjustments based on the information of previous or current prompts to determine the best direction for improving them.
We apply the descent direction and momentum concepts to design the meta-prompts.

\subsubsection{Analogical Descent Direction.}
Descent direction determines the direction of parameter updates based on the model performance.
{We analogize two similar forms that determine how to improve new prompts according to the current information.}

\begin{table*}[t]
    \centering
    \tiny
    \resizebox{0.9\textwidth}{!}{%
    \begin{tabular}{cl|cc|cc|cc}
        \toprule
        \multicolumn{2}{c|}{\textbf{Prompt Optimizer}} & \multicolumn{2}{c|}{{GPT-3.5-turbo}} & \multicolumn{2}{c|}{GPT-4} & \multicolumn{2}{c}{GPT-4} \\
        \midrule
        \multicolumn{2}{c|}{\textbf{Task Model}} & \multicolumn{2}{c|}{Llama-2-7b-chat} & \multicolumn{2}{c|}{Llama-2-7b-chat} & \multicolumn{2}{c}{GPT-3.5-turbo} \\
        \midrule
        \multicolumn{2}{c|}{\textbf{Gradient}} & P+M & P+M+R & P+M & P+M+R & P+M & P+M+R \\
        \midrule
        \multirow{5}{*}{\textbf{Momentum}} & None & 41.07 & 40.34 & 40.32 & 39.56 & 64.76 & 64.55 \\
        & Summarization                                & 41.03 & 40.63 & 40.58 & 40.41 & 64.62 & 64.55 \\
        & Recency                              & 41.93 & 41.55 & 42.02 & 41.34 & 65.26 & 64.97 \\
        & Relevance                                                  & \textbf{42.53} & 40.81 & \textbf{42.89} & 41.97 & \textbf{65.87} & 65.39 \\
        & Importance                                                 & 41.84 & 39.73 & 41.73 & 41.04 & 65.26 & 65.11 \\
        \bottomrule
    \end{tabular}
    }
    \caption{The performance comparison of various update directions based on different prompt optimizers and task models. ``P'' denotes prompt, ``M'' denotes performance, and ``R'' denotes reflection.}
    \label{tab:source_exp}
\end{table*}

$\bullet$ \emph{Prompt+Performance.}
One straightforward method is to include the last-round task prompt and the corresponding model performance into the meta-prompt for the LLM-based optimizer $\mathcal{M}_O$. 
It leverages the capacity of LLMs to reason about how to improve prompting optimization. 

$\bullet$ \emph{Prompt+Performance+Reflection.}
Another way to solve the barrier of the descent direction is to leverage the reflection capability of LLMs~\cite{Automatic-Pryzant-2023-EMNLP}.
With the reflection mechanism, LLMs can generate feedback from past failures to improve performance.
Such feedback can be seen as a form of {``semantic'' gradient signals.} 

\subsubsection{Analogical Momentum.} 
Inspired by the momentum in gradient descent, 
{we aim to make LLMs aware of the prompts used in the optimization process and their corresponding results (\ie descent direction), thereby achieving a more global update direction.}
A straightforward way is to directly include all {intermediate results} into the meta-prompt. 
However, it might be limited by the context length of LLMs and affected by the accumulated noise.
To better utilize the optimization trajectory, we propose two alternative methods.

$\bullet$ \emph{Summarization-based trajectory.}
One simple approach is to summarize the {intermediate results} from the optimization trajectory.
Specifically, at each step, we use an instruction {(\ie ``{Your task is to summarize the problems in the previous prompt and the current prompt.}'')} to let the LLM perform summarization using the summary in the last step and the result in the current step.

 $\bullet$ \emph{Retrieval-based trajectory.}
Another approach is to select $k$ pieces of intermediate results from the optimization trajectory.
Specifically, we consider three strategies:
(1) Recency: selecting $k$ nearest intermediate results;
(2) Relevance: selecting $k$ most relevant intermediate results, which are measured by the semantic similarity based on the BGE model~\cite{C-Pack-Xiao-2023-arXiv}; 
(3) Importance: selecting $k$ most important intermediate results, which are measured by the performance.

\subsection{Update Method}
\label{subsec:application}
{The update method can refer to how the LLM utilizes such information to improve task prompts.}
Accordingly, we explore how to analogize the learning rate and the descent process into the design of meta-prompts. 

\subsubsection{Analogical Learning Rate.}
In the model optimization, the learning rate controls the extent of gradient updates at each step. 
Similarly, we aim to control the variation degree of prompt optimization. 
Specifically, we limit the number of words that can be modified {in the meta-prompt} (\ie \emph{edit distance}).
Accordingly, we propose two methods of controlling edit distance as follows: 

$\bullet$ \emph{Decay-based constraint.}
To avoid overshooting the optimal solution, the decay strategy is proposed to gradually reduce the learning rate~\cite{Averaging-Izmailov-2018-UAI}.
Here, we borrow the idea of controlling the maximum edit distance and consider gradually reducing its value following either a linear or cosine curve.
In particular, we reduce the constraint to approximately {20\%} of its maximum value until convergence.

$\bullet$ \emph{Warmup-based constraint.}
To avoid instability in the early stage of optimization, the warmup strategy is proposed to gradually increase the learning rate at the beginning~\cite{Accurate-Goyal-2017-arXiv}.
Similarly, we adopt the widely used linear warmup schedule to gradually increase the constraint for the maximum edit distance in the initial {5\%} steps.

\subsubsection{Analogical Gradient Descent.}
By analogy with the subtraction operation in gradient descent (\ie $-\tau_k g_k$ in Eq.~\eqref{eq:update}), we introduce two methods to update the task prompt accordingly.

$\bullet$ \emph{Editing-based refinement.}
The first method directly edits the last-round task prompt to improve performance.
Specifically, we add an instruction {(\ie ``{Modify the current prompt}'')} into the meta-prompt, which requires the LLM to edit the task prompt from the last step according to the update direction.
This method allows for effectively exploiting the existing prompt.

$\bullet$ \emph{Generation-based refinement.}
In addition to direct edits, we can also leverage the in-context learning capability of LLMs to generate refined task prompts.
Specifically, we present the information regarding the updated direction in the format of demonstrations.
Then, We add an instruction (\ie ``Write a new improved prompt'') to let the LLM follow the demonstration to directly generate a new task prompt.
Compared with the editing-based method, the generation-based approach explores a wider range of prompt variations.

\begin{table*}[ht!]
    \centering
    \tiny
    \resizebox{0.95\textwidth}{!}{%
    \begin{tabular}{cl|cc|cc|cc}
        \toprule
        \multicolumn{2}{c|}{\textbf{Prompt Optimizer}} & \multicolumn{2}{c|}{GPT-3.5-turbo} & \multicolumn{2}{c|}{GPT-4} & \multicolumn{2}{c}{GPT-4} \\
        \midrule
        \multicolumn{2}{c|}{\textbf{Task Model}} & \multicolumn{2}{c|}{Llama-2-7b-chat} & \multicolumn{2}{c|}{Llama-2-7b-chat} & \multicolumn{2}{c}{GPT-3.5-turbo} \\
        \midrule
        % \multicolumn{2}{c|}{\textbf{Initial Prompt}} & \multicolumn{2}{c|}{Let's think step by step: 31.25} & \multicolumn{2}{c|}{Let's think step by step: 31.25} & \multicolumn{2}{c}{Let's think step by step: 62.15} \\
        % \midrule
        \multicolumn{2}{c|}{\textbf{Learning rate}} & Editing & Generation & Editing & Generation & Editing & Generation \\
        \midrule
        \multirow{7}{*}{\textbf{Descent}} & No control & 42.53 & 42.61 & 42.89 & 43.17 & 65.87 & 66.37 \\
        % \cmidrule(lr){2-2}
        & Fixed & 42.91 & 43.09 & 43.38 & 43.66 & 66.48 & 66.91 \\
        & +Warmup & 41.76 & 40.08 & 42.53 & 42.95 & 65.79 & 65.52 \\
        % \cmidrule(lr){2-2}
        & Linear decay & 42.68 & 42.86 & 43.55 & 44.03 & 66.56 & 67.10 \\
        & +Warmup & 41.47 & 41.12 & 41.47 & 42.91 & 66.03 & 66.18 \\
        % \cmidrule(lr){2-2}
        & Cosine decay & 42.95 & \textbf{43.75} & 43.98 & \textbf{44.97} & 66.74 & \textbf{67.80} \\
        & +Warmup & 40.19 & 41.29 & 42.68 & 43.13 & 65.94 & 66.37 \\
        \bottomrule
    \end{tabular}
    }
    \caption{The performance comparison of various update methods based on different prompt optimizers and task models.}
    \label{tab:application_exp}
\end{table*}

\subsection{Empirical Experiments}
\label{subsec:study-exp}
In this part, we describe the experiment setting for our analogical analysis and detail our findings from the experiment.

\subsubsection{Experiment Setup.}
\label{sec:study-exp-setup}
We select a dataset from each type of task in BBH~\cite{Challenging-Suzgun-2023-ACL} to create a lite BBH benchmark for our analysis:
i) Navigate (binary choice);
ii) Movie Recommendation (multiple choice);
iii) Object Counting (numeric response);
iv) Word Sorting (free response).
We adopt exact match as the metric for performance evaluation.
{We use three different model combinations of prompt optimizers and task models  (\ie  \texttt{GPT-3.5-turbo} and \texttt{Llama-2-7b-chat}, \texttt{GPT-4} and \texttt{Llama-2-7b-chat}, \texttt{GPT-4} and \texttt{GPT-3.5-turbo}).
{The optimization process lasts for at most 3 epochs, under which the task prompt can reach the plateau}.

\subsubsection{Findings for Update Direction.}
The results for the update direction are presented in Table~\ref{tab:source_exp}.
Here are the main findings:

$\bullet$ \emph{Reflection Leads to Performance Drop.} 
Regarding the analogy to descent direction, \textit{prompt+performance} achieves better performance than \textit{prompt+performance+reflection}, which brings an improvement of up to 31\% compared with the initial task prompt. 
The improvement brought by prompts can be attributed to their rich semantic information about the task, which can activate the task-specific knowledge of LLMs for optimization.
In contrast, LLMs are limited in their capabilities of reflection~\cite{Large-Huang-2023-arXiv}, which may lead to inaccurate updates.

$\bullet$ \emph{Prompt Optimizers can Learn More from Contextually Relevant Prompts.} 
The analogical concept of momentum can further improve the performance.
Among various designs, \textit{relevance-based trajectory} emerges as the most effective one, which brings an additional 15\% improvement. 
This can be attributed to the fact that LLMs can learn more from contextually relevant prompts, while it might be challenging for LLMs to fully understand the signal of recency or importance.
By contrast, the summarization-based trajectory proves to be less helpful.
One possible reason is that summarization only captures common aspects of the trajectory while neglecting details that may be crucial.

\subsubsection{Findings for Update Method.}
To investigate the update method for prompt optimization, we conduct experiments using the best configuration found in the previous experiments.
The results for the update method are presented in Table~\ref{tab:application_exp}.

$\bullet$ \emph{Generation-based Refinement is Better.} 
In general, \textit{generation-based refinement} outperforms \textit{editing-based refinement}, which brings an improvement of up to 36\% compared with the initial task prompt. 
This can be partly attributed to the significance of exploration in prompt optimization.
Generation-based strategy is not confined to the current prompt, allowing the model to better leverage the in-context examples. 
Therefore, this approach can demonstrate great flexibility to enable LLMs to explore a larger search space.

$\bullet$ \emph{Decay Strategy is Helpful.}
Among various controlling methods for prompt variation, \textit{cosine decay-based constraint} achieves the best performance, bringing an additional 10\% improvement.
However, unlike gradient-based model optimization, the warmup strategy does not yield improvement.
This finding suggests that, at the early stage of prompt optimization, exploration plays a crucial role in conducting large-scale prompt searches, while stability becomes more important in the later stage for more refined adjustments.

\section{Method}
\label{sec:OURS}

\begin{table}[t]
\centering
% \large
\resizebox{\linewidth}{!}{%
\begin{tabular}{c|lc|cc}
\toprule
\multirow{3}{*}{\begin{tabular}[c]{@{}c@{}}\textbf{Prompt}\\\textbf{optimizer}\end{tabular}}                    & \multicolumn{2}{c|}{\textbf{Update direction}}                  & \multicolumn{2}{c}{\textbf{Update method}}                   \\
\cmidrule{2-5}
& \multicolumn{1}{c}{\begin{tabular}[c]{@{}c@{}}Currect\\information\end{tabular}}   & \begin{tabular}[c]{@{}c@{}}Trajectory\end{tabular}           & \begin{tabular}[c]{@{}c@{}}Edit\\distance\end{tabular} & \begin{tabular}[c]{@{}c@{}}Refinement\\strategy\end{tabular}    \\
% \midrule
% \multirow{2}{*}{\begin{tabular}[c]{@{}l@{}}Model\\Optimizer\end{tabular}}     & SGD  & Gradient & X     &                          &          \\
                                                                              % & SGDM & Prompt+performance+reflection    & Summarization-based trajectory   & No     & Editing-based refinement         \\
\midrule
% APE~\cite{Large-Zhou-2023-ICLR}             & P         & None                          & None                      & G        \\
% APO~\cite{Automatic-Pryzant-2023-EMNLP}     & P+R       & None                          & None                      & E        \\
% OPRO~\cite{Large-Yang-2023-arXiv}           & P+M       & Recency                       & None                      & G        \\
% PE2~\cite{Prompt-Ye-2023-arXiv}             & P+M+R     & Recency                       & Fixed                     & G        \\
APE             & P         & None                          & None                      & Generation        \\
APO             & P+R       & None                          & None                      & Editing           \\
OPRO            & P+M       & Recency                       & None                      & Generation        \\
PE2             & P+M+R     & Recency                       & Fixed                     & Generation        \\
\midrule                                            
{\OURS}         & P+M       & Relevance                      & Decay                     & Generation        \\
\bottomrule
\end{tabular}%
}
\caption{Comparisons of \OURS with existing LLM-based prompt optimizers.
% including APE~\cite{Large-Zhou-2023-ICLR}, APO~\cite{Automatic-Pryzant-2023-EMNLP}, OPRO~\cite{Large-Yang-2023-arXiv}, and PE2~\cite{Prompt-Ye-2023-arXiv}. 
``P'' refers to prompt, ``M'' refers to performance, and ``R'' refers to reflection.}
\label{tab:comparison}
\end{table}

\begin{table*}[ht!]
\centering
\tiny
\resizebox{0.95\textwidth}{!}{%
\begin{tabular}{c|cc|ccccc|cc}
\toprule
\multirow{2}{*}{\centering \textbf{Task}}     & \multicolumn{2}{c|}{\multirow{2}{*}{\begin{tabular}[c]{@{}c@{}}\textbf{Complex}\\\textbf{reasoning task}\end{tabular}}} & \multicolumn{5}{c|}{\multirow{2}{*}{\centering \textbf{Knowledge intensive task}}} & \multicolumn{2}{c}{\multirow{2}{*}{\begin{tabular}[c]{@{}c@{}}\textbf{Common}\\\textbf{NLP task}\end{tabular}}} \\
         &                     &        & & & &             &              &                        &                          \\
\midrule
\multirow{2.5}{*}{\centering Dataset} & \multirow{2.5}{*}{\centering BBH}               & \multirow{2.5}{*}{GSM8K}             & \multicolumn{5}{c|}{MMLU}       & \multirow{2.5}{*}{\centering WSC}                  & \multirow{2.5}{*}{\centering WebNLG}  \\                
\cmidrule{4-8}
                         &                                    &                                    & STEM & Human. & Social. & Other & Average & & \\
\midrule
Empty                                          &  30.51              &  22.00     & 31.05 & 36.54 & 41.75 & 37.20     &  35.96       & 60.67                  & 32.14                     \\
{\begin{tabular}[c]{@{}c@{}}CoT\end{tabular}}  &  29.91              &  24.00     & 32.53 & 37.05 & 41.05 & 36.94     &  36.36       & 59.33                  & 31.11                     \\ 
\midrule
SGDM           &  33.30              & 27.33      & 32.88 & 38.36 & 41.88 & 38.02 & 37.20              & 64.00                  & 38.01                     \\
APE            &  32.94              & 25.00      & 33.51 & 38.69 & 42.02 & 37.96 & 37.50              & 62.00                  & 36.49                     \\
APO            &  32.97              & 25.33      & 33.17 & 37.94 & 44.94 & 38.23 & 37.71              & 62.00                  & 34.92                     \\
OPRO           &  33.29              & 26.67      & 34.76 & 38.72 & 43.55 & 37.11 & 38.05              & 63.33                  & 37.89                     \\
PE2            &  33.43              & 25.33      & 33.77 & 37.95 & 44.80 & 38.25 & 38.07              & 62.67                  & 39.10                     \\
\midrule
\OURS          & \textbf{35.43}      & \textbf{28.33} &  \textbf{35.00} & \textbf{38.84} & \textbf{46.61} & \textbf{38.60} & \textbf{39.14}     & \textbf{65.33}         & \textbf{42.51}                     \\
\bottomrule
\end{tabular}%
}
\caption{
Performance comparison using only the task prompts obtained from different methods. ``Human.'' and ``Social.'' denote the datasets classified as Humanities and Social Science in MMLU.
}
\label{tab:experiment}
\end{table*}

{In this section, we present our novel gradient-inspired LLM-based prompt optimizer called \textbf{\OURS}}, which leverages the insights gained from our systematic study.
\OURS adopts an iterative prompt optimization framework.
At each step, it first retrieves relevant prompts from the optimization trajectory as the update direction.
Then, it utilizes the generation-based refinement strategy to perform the update, while controlling the edit distance through a cosine-based decay strategy.

\subsubsection{Iterative Prompt Optimization.}
\OURS performs prompt optimization through a multi-step iterative process. 
At each step, it first generates multiple candidate task prompts based on a meta-prompt. 
The meta-prompt serves as the input that guides the LLM in optimizing its prompts. 
Subsequently, we select the task prompt with the best performance for the next iteration.
This iterative process continues until either the performance of the task prompt reaches a plateau or the predefined maximum number of optimization steps is reached.

\subsubsection{The Design of Meta-Prompt.}
As the input to the LLM, our meta-prompt consists of two key components: update direction and update method.

$\bullet$ \emph{Update direction.}
To determine the update direction, we leverage the retrieval-based optimization trajectory.  
This trajectory consists of past task prompts, along with their model performance. 
They are selected using a \textit{relevance-based strategy} and sorted in ascending order based on their similarity to the current prompt.

$\bullet$ \emph{Update method.}
After the update direction is determined, we further employ the \textit{generation-based refinement} strategy to update the task prompt. 
Specifically, we present the trajectory in the format of demonstrations in the meta-prompt.
Then, the LLM can follow these demonstrations to generate a new task prompt via in-context learning.
Additionally, we implement the \textit{cosine-based decay strategy} to control the edit distance between task prompts at consecutive iterations, ensuring gradual and controllable changes.

Furthermore, we enhance the meta-prompt by incorporating a few task examples.
These examples provide additional context to aid the LLM in understanding the task effectively.
% {The complete meta-prompt is presented in Appendix F.}

\subsubsection{Comparison of LLM-Based Prompt Optimizers.}
Existing LLM-based prompt optimizers can be divided into two main classes by their update direction.
Some work (\ie APO~\cite{Automatic-Pryzant-2023-EMNLP} and PE2~\cite{Prompt-Ye-2023-arXiv}) leverages the reflection capability of LLMs to produce textual ``gradients'' as the update direction, while others (\ie OPRO~\cite{Large-Yang-2023-arXiv} and APE~\cite{Large-Zhou-2023-ICLR}) directly use task prompts as the update direction.
Our approach is based on the systematic investigation of the update direction and the update method.
In particular, we propose several novel designs for the meta-prompt: relevance-based trajectory as the update direction and decay-based constraint for edit distance in the update method.
Table~\ref{tab:comparison} presents a detailed comparison. 

\section{Experiments}
\label{sec:exp}

In this section, we first set up the experiments and then report the results and detailed analysis.

\subsection{Experimental Setup}
\label{subsec:experimental_setup}

\subsubsection{Tasks and Datasets.}
We select datasets from three groups of tasks for the experiment: Big-Bench Hard~(BBH)~\cite{Challenging-Suzgun-2023-ACL} and GSM8K~\cite{Training-Cobbe-2021-arXiv} for complex reasoning tasks, MMLU~\cite{Measuring-Hendrycks-2021-arXiv} for knowledge-intensive tasks, and WSC~\cite{The-Levesque-2012-KR} and WebNLG~\cite{Creating-Gardent-2017-ACL} for common NLP tasks.
Due to computational limitations, we sample a subset from each dataset for the main experiment.
In addition, we use the lite BBH benchmark for detailed analysis.
% {Other details are presented in Appendix A.2.}

\subsubsection{Baselines.}
We select several representative methods for comparison, including existing LLM-based prompt optimizers and one from gradient-based model optimizers.
(1) \underline{\emph{SGDM}}~\cite{On-Sutskever-2013-ICML} is a momentum-based model optimizer.
We adapt it for prompt optimization using the summarization-based trajectory and the editing-based refinement strategy.
(2) \underline{\emph{APE}}~\cite{Large-Zhou-2023-ICLR} utilizes LLMs to generate semantically similar variants of task prompts and selects one with the best performance.
(3) \underline{\emph{APO}}~\cite{Automatic-Pryzant-2023-EMNLP} first uses reflection to obtain the gradient and then edits the task prompt accordingly.
(4) \underline{\emph{OPRO}}~\cite{Large-Yang-2023-arXiv} incorporates historical prompts with their scores into the meta-prompt.
(5) \underline{\emph{PE2}}~\cite{Prompt-Ye-2023-arXiv} adds historical prompts and reflection into the meta-prompt and controls the edit distance with a fixed constraint.
In addition, we consider the baseline without an instruction~(``Empty'') and the instruction ``Let' think step by step.'' from chain-of-thought prompting~\cite{Large-Kojima-2022-NeurIPS} for performance comparison.

\begin{table}[t]
\centering
\resizebox{\columnwidth}{!}{%
\begin{tabular}{c|cc|cccc}
\toprule
\textbf{Task model} & \begin{tabular}[c]{@{}c@{}}\textbf{Llama-}\\ \textbf{2-7b}\end{tabular} & \begin{tabular}[c]{@{}c@{}}\textbf{Llama-}\\ \textbf{3-8b}\end{tabular} & \multicolumn{2}{c}{\begin{tabular}[c]{@{}c@{}}\textbf{Llama-2-}\\ \textbf{7b-chat}\end{tabular}} & \multicolumn{2}{c}{\begin{tabular}[c]{@{}c@{}}\textbf{Llama-3-}\\ \textbf{8b-instruct}\end{tabular}} \\
\midrule
\textbf{Setting} & \multicolumn{2}{c|}{\begin{tabular}[c]{@{}c@{}}\textbf{I + D}\end{tabular}} & \begin{tabular}[c]{@{}c@{}}\textbf{I}\end{tabular} & \begin{tabular}[c]{@{}c@{}}\textbf{I + D}\end{tabular} & \begin{tabular}[c]{@{}c@{}}\textbf{I}\end{tabular} & \begin{tabular}[c]{@{}c@{}}\textbf{I + D}\end{tabular} \\
\midrule
Empty       & 40.28 & 60.42 & 32.29   & 36.63 & 48.26 & 54.86 \\
CoT         & 36.46 & 51.39 & 31.25   & 34.20 & 52.43 & 54.34 \\
\midrule
SGDM        & 42.19  & 62.50 & 40.63  & 35.77  & 60.42 & 58.51 \\
APE         & 42.54  & 61.28 & 42.01  & 36.29  & 59.43 & 58.51 \\
APO         & 42.19  & 61.45 & 40.34  & 36.29  & 59.38 & 57.99 \\
OPRO        & 42.02  & 62.68 & 42.14  & 36.46  & 62.33 & 59.02 \\
PE2         & 42.88  & 63.20 & 42.01  & 36.81  & 62.67 & 58.68 \\
\midrule
\OURS   & \textbf{45.48} & \textbf{65.11} & \textbf{43.75} & \textbf{38.02} & \textbf{63.89}  & \textbf{61.11} \\ 
\bottomrule
\end{tabular}
}
\caption{Performance comparison under different evaluation settings. ``I'' denotes instruction and ``D'' denotes demonstration.}
\label{tab:exp-few-shot}
\end{table}

\begin{table*}[t]
    \centering
    % \large
    \resizebox{\textwidth}{!}{%
    \begin{tabular}{c|ccccc|ccccc}
        \toprule
        \begin{tabular}[c]{@{}c@{}}\textbf{Prompt}\\\textbf{optimizer}\end{tabular} & \multicolumn{5}{c|}{GPT-3.5-turbo} & \multicolumn{5}{c}{GPT-4} \\
        \midrule
        \begin{tabular}[c]{@{}c@{}}\textbf{Task}\\\textbf{model}\end{tabular} & \begin{tabular}[c]{@{}c@{}}Baichuan2-\\7b-chat\end{tabular} & \begin{tabular}[c]{@{}c@{}}Llama-2-\\7b-chat\end{tabular} & \begin{tabular}[c]{@{}c@{}}Llama-2-\\13b-chat\end{tabular} & \begin{tabular}[c]{@{}c@{}}GPT-3.5-\\turbo\end{tabular} & GPT-4 & \begin{tabular}[c]{@{}c@{}}Baichuan2-\\7b-chat\end{tabular} & \begin{tabular}[c]{@{}c@{}}Llama-2-\\7b-chat\end{tabular} & \begin{tabular}[c]{@{}c@{}}Llama-2-\\13b-chat\end{tabular} & \begin{tabular}[c]{@{}c@{}}GPT-3.5-\\turbo\end{tabular} & GPT-4 \\
        \midrule
        Empty & 15.75 & 32.29 & 40.97 & 60.48 & 72.87 & 15.75 & 32.29 & 40.97 & 60.48 & 72.87 \\
        % \midrule
        CoT & 19.45 & 31.25 & 40.28 & 62.15 & 73.61 & 19.45 & 31.25 & 40.28 & 62.15 & 73.61 \\
        \midrule
        SGDM & 20.87 & 40.63 & 42.24 & 64.18 & 75.23 & 21.85 & 40.41 & 42.03 & 64.55 & 75.82 \\
        % \midrule
        APE & 20.61 & 42.01 & 41.65 & 63.63 & 74.71 & 20.29 & 39.98 & 42.96 & 64.38 & 74.13 \\
        % \midrule
        APO & 19.97 & 40.34 & 41.79 & 63.96 & 74.13 & 22.09 & 39.56 & 43.67 & 64.55 & 74.56 \\
        % \midrule
        OPRO & 19.86 & 42.14 & 42.89 & 64.56 & 75.06 & 21.28 & 41.77 & 43.35 & 64.91 & 76.67 \\
        % \midrule
        PE2 & 21.11 & 42.01 & 42.68 & 64.95 & 75.27 & 22.43 & 42.03 & 44.91 & 65.23 & 76.55 \\
        \midrule
        \OURS & \textbf{23.35} & \textbf{43.75} & \textbf{44.83} & \textbf{67.02} & \textbf{76.56} & \textbf{25.34} & \textbf{44.97} & \textbf{46.17} & \textbf{67.80} & \textbf{78.65} \\
        \bottomrule
    \end{tabular}
    }
    \caption{Performance of different prompt optimization methods with various models.}
    \label{tab:model_selection}
\end{table*}

% \begin{table}[t]
% \centering
% \small
% \resizebox{\columnwidth}{!}{%
% \begin{tabular}{cc|c}
% \toprule
% \begin{tabular}[c]{@{}c@{}}\textbf{Prompt} \textbf{optimizer}\end{tabular}               & \begin{tabular}[c]{@{}c@{}}\textbf{Task} \textbf{model}\end{tabular} & \begin{tabular}[c]{@{}c@{}}\textbf{Accuracy}\\\textbf{(before / after)}\end{tabular} \\
% \midrule
% \multirow{3}{*}{GPT-3.5-turbo}      & Llama-2-7b-chat   & 31.25 / 43.75 \\
%                                     & GPT-3.5-turbo     & 62.15 / 67.02 \\
%                                     & GPT-4             & 73.61 / 76.56 \\
% \midrule
% \multirow{3}{*}{GPT-4}              & Llama-2-7b-chat   & 31.25 / 44.97 \\
%                                     & GPT-3.5-turbo     & 62.15 / 67.80 \\
%                                     & GPT-4             & 73.61 / 78.65 \\
% \bottomrule
% \end{tabular}%
% }
% \caption{The performance before/after prompt optimization with different models as the prompt optimizer and the task model.}
% \label{tab:model_selection}
% \end{table}

\subsubsection{Evaluation Metrics.}
We report the average accuracy of all the subtasks for BBH and MMLU following \citet{Challenging-Suzgun-2023-ACL} and \citet{Measuring-Hendrycks-2021-arXiv}, accuracy for GSM8K following \citet{Training-Cobbe-2021-arXiv}, ROUGE-L for WSC and WebNLG following \citet{Super-Wang-2022-EMNLP}.

\subsubsection{Implementation Details.}
We use both the base model (\ie \texttt{Llama-2-7b} and \texttt{Llama-3-8b}) and the instruction-tuned models (\ie \texttt{Baichuan2-7b-chat}, \texttt{Llama-2-7b-chat}, \texttt{Llama-2-13b-chat}, \texttt{Llama-3-8b-instruct}~\cite{llama3-8b}, \texttt{GPT-3.5-turbo}, and \texttt{GPT-4}) as the task model.
For the prompt optimizer, we use \texttt{GPT-3.5-turbo} and \texttt{GPT-4}.
Unless otherwise specified, we use \texttt{Llama-2-7b-chat} as task model and \texttt{GPT-3.5-turbo} as prompt optimizer throughout experiments.
We repeat all the experiments three times and report the average of the results.
% {Other details are presented in Appendix A.2.}

\subsection{Main Results}
Table~\ref{tab:experiment} and~\ref{tab:exp-few-shot} show the results of different methods for prompt optimization across various tasks and evaluation settings.
% {Detailed experiment results are shown in Appendix E.}

First, when only considering the task prompt, we can see that 
trajectory-based methods (\ie SGDM, OPRO, PE2, and \OURS) perform very well.
One possible reason is that the trajectory helps the prompt optimizer pay more attention to the important information instead of the noise in the current step.
Furthermore, our prompt optimizer \OURS achieves the best performance across all tasks.
Our relevance-based trajectory provides semantically similar demonstrations that can be effectively learned by the LLM, while the cosine-based decay strategy can control the optimization process in a fine-grained manner through edit distance.

Second, under various evaluation settings for the lite BBH benchmark, it can be observed that \OURS not only excels in the ``Instruction'' setting but also yields substantial gains in the ``Instruction + Demonstration'' setting for both the base model and the instruction-tuned variant.
Even in the scenario that is challenging for baselines (\ie \texttt{Llama-2-7b-chat} with ``Instruction + Demonstration''), our approach still demonstrates strong improvement.
This showcases the versatility of our approach in both zero-shot and few-shot evaluation settings.

\subsection{Detailed Analysis}
Next, we conduct a detailed analysis of our prompt optimizer \OURS from the following aspects.
% {Other detailed analysis are presented in Appendix B}.

\subsubsection{The Impact of Model Selection.}
To confirm the effectiveness of \OURS across different models, we explore the impact of different model combinations compared with other baseline methods.
Table~\ref{tab:model_selection} presents the results on the lite BBH benchmark.
In general, \OURS demonstrates remarkable capabilities for prompt optimization across various scenarios, including strong-to-weak optimization, self-optimization, and weak-to-strong optimization.
This indicates the versatility of our framework.
{Notably, \texttt{GPT-4} can consistently find better task prompts than \texttt{GPT-3.5-turbo}, which suggests the need for a capable model as the prompt optimizer.}

\subsubsection{The Efficiency of Optimization.}
\label{para:efficency}

\begin{figure}[t]
    \centering
    \begin{subfigure}[b]{0.49\linewidth}
        \centering
        \includegraphics[width=\textwidth]{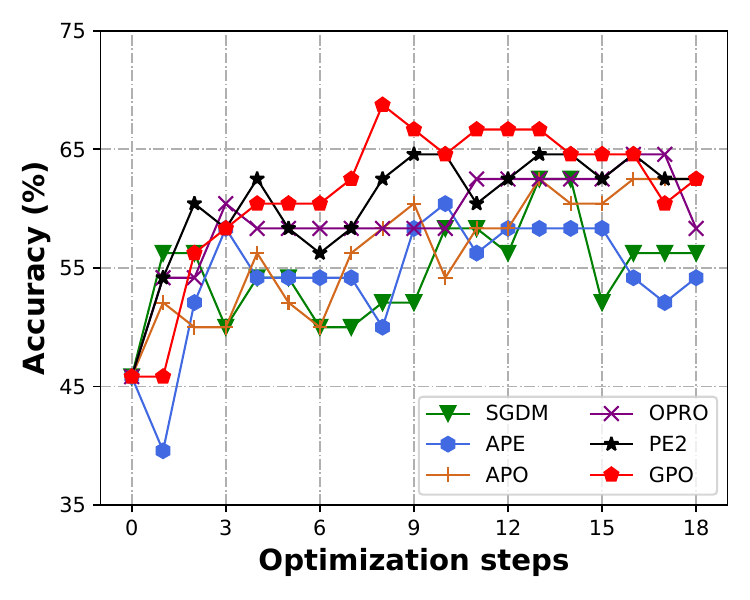}
        \caption{Optimization curve}
        \label{fig:optimization_curve}
    \end{subfigure}
    \begin{subfigure}[b]{0.49\linewidth}
        \centering
        \includegraphics[width=\textwidth]{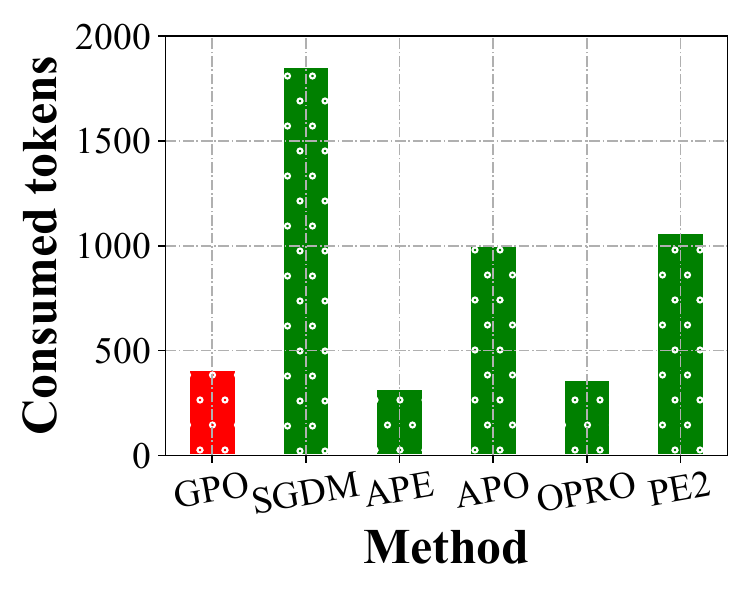}
        \caption{Token consumption}
        \label{fig:consumed_tokens}
    \end{subfigure}
    \caption{The efficiency of our approach \OURS w.r.t. optimization steps and token consumption.}
\label{fig:efficiency}
\end{figure}

% \begin{figure}[t]
%     \centering
%     \centering
%     \includegraphics[width=0.9 \linewidth]{figures/optimization_steps.pdf}
%     \caption{The efficiency of our approach \OURS w.r.t. optimization steps.}
% \label{fig:efficiency}
% \end{figure}

LLM-based prompt optimization requires multiple interactions with the LLM.
% In this part, we investigate the efficiency of \OURS from the optimization curve of the first 12 steps and token consumption.
In this part, we investigate the efficiency of LLM-based prompt optimizers by examining the optimization curve over the first 12 steps.
% The results are presented in Figure~\ref{fig:efficiency}.
Figure~\ref{fig:efficiency} shows that on the movie recommendation dataset, compared to other methods, \OURS demonstrates rapid enhancement of the task prompt in the early stage, followed by steady and consistent performance improvement in the later stage of optimization.
% {Optimization curves on other datasets are detailed in Appendix B.1.}
% Second, as depicted in Figure~\ref{fig:consumed_tokens}, the average token consumption of \OURS on the lite BBH benchmark is much lower than SGDM, APO, and PE2, and comparable to APE and OPRO.
Since \OURS only utilizes task prompts to derive the update direction and performs fine-grained control over the variation, it can achieve better performance with high efficiency.

\section{Conclusion and Discussion}
% \textcolor{blue}{
In this paper, we conduct a systematic analogy between gradient-based model optimizers and LLM-based prompt optimizers.
% Based on such an analogy framework, we conduct extensive studies and find some empirical results.
Based on existing work and our newly proposed approach, we conduct an experimental analysis of the two key factors (\ie update direction and update method) to determine the best configuration.
According to this configuration, we further propose a novel prompt optimization framework~{\OURS}.
At each step, it retrieves relevant prompts from the optimization trajectory as the update direction.
Then, it utilizes the generation-based refinement strategy to perform the update, while controlling the edit distance through a cosine-based decay strategy.
Extensive experiments demonstrate the effectiveness and efficiency of \OURS.

One limitation of our work is that we only draw an analogy with the most widely used gradient-based optimizers.
More advanced model optimizers like Newton's method and its application to meta-prompts remain to be investigated.
Additionally, our approach relies on textual update directions, future research could explore more direct and fine-grained numerical update signal methods~\cite{Nie-Numerical-Optimizer}.

\section*{Acknowledgements}
This work was partially supported by National Natural Science Foundation of China under Grant No. 92470205 and 62222215. Xin Zhao is the corresponding author.

\bibliography{aaai25}

\section{A  Additional Details for the Experimental Setup}

In this part, we detail the setup for the analogical analysis and experiment.

\subsection{A.1 Additional Details for the Setup of Analogical Analysis}
\label{app:setup-for-analogical-analysis}

\paratitle{Tasks and Metrics.}
Following~\citet{Large-Yang-2023-arXiv}, we utilize Big-Bench Hard~(BBH)~\cite{Challenging-Suzgun-2023-ACL} for evaluation.
BBH is a suite of 23 challenging BIG-Bench tasks~\cite{Beyond-Srivastava-2022-arXiv} that covers various kinds of reasoning capabilities.
Due to the constraints of computational resources, we select a dataset from each type of task in BBH to create a lite benchmark for our analysis:
i) Navigate (Binary choice);
ii) Movie Recommendation (Multiple choice);
iii) Object Counting (Numeric response);
iv) Word Sorting (Free response).
For each dataset, we split it into train/valid/test sets with a ratio of 2:2:6.
Following~\citet{Challenging-Suzgun-2023-ACL}, we adopt the exact match as the metric for performance evaluation.

% \paratitle{Baseline Prompt Optimizer.}
% To effectively reflect the contribution of each factor, we construct a simple LLM-based prompt optimizer as the baseline.
% Specifically, we utilize ChatGPT as the backbone for the prompt optimizer due to its powerful capability of instruction following.
% For simplicity, we take the prompt itself as the derivation of information and utilize it by editing the current prompt without controlling the edit distance.
% To implement the above designs, we craft a prompt to help ChatGPT understand and follow them, which is usually called \textit{meta-prompt}~\cite{Large-Yang-2023-arXiv}.
% In each optimization step, we first instruct ChatGPT with the meta-prompt to generate candidate prompts, and then select the best-performing one as the prompt for the next iteration following~\citet{Large-Yang-2023-arXiv, Prompt-Ye-2023-arXiv}.

\paratitle{Implementation Details.}
Following~\citet{Agent-Crispino-2023-arXiv}, we select \texttt{Llama-2-7b-chat} as the task model and set its temperature to 0 to make the output as deterministic as possible.
For the prompt optimizer, we employ \texttt{GPT-3.5-turbo}, which is the underlying model of ChatGPT.
We set its temperature to 1.0 to encourage the generation to be more diverse.
To help the prompt optimizer understand the downstream task, following~\citet{Large-Yang-2023-arXiv}, we randomly sample 3 examples from the dataset and fill them into the meta-prompt of the prompt optimizer.
In the process of optimization, we take the instruction ``Let's think step by step.'' as the initial prompt and insert it at the end of the question to obtain better performance following~\citet{Challenging-Suzgun-2023-ACL}.
% which is widely used for chain-of-thought prompting and obtains great improvement in complex reasoning tasks~\cite{}.
At each step, we first prompt the optimizer to generate 8 candidate task prompts, and then select the best-performing one as the task prompt for the next iteration following~\citet{Large-Yang-2023-arXiv, Prompt-Ye-2023-arXiv}.
The optimization process lasts for at most 3 epochs with a batch size of 8.
% \textcolor{blue}{Early stopping is used with a patience of 5 steps.}
We repeat all the experiments three times and report the average of the results.
The meta-prompts we used are detailed in Appendix~\ref{app:meta-prompt_analysis}.

\subsection{A.2 Additional Details for the Setup of Experiment}
\label{app:dataset-for-exp}

\subsection{The Statistics of Datasets}
\label{app:Dataset_Statistic}
As mentioned in the main body, we sample a subset of the dataset for efficient evaluation.
For BBH and MMLU, we split the entire dataset into training, validation, and test sets with a ratio of 2:2:6.
For GSM8k and WebNLG, we randomly sample 100 examples as the training set, 100 for the validation set, and 300 for the test set.
For WSC, we randomly sample 50 examples as the training set, 50 for the validation set, and 150 for the test set.

\subsection{Implementation Details}
\label{app:exp-details}
For the task model, we use both the base model (\ie \texttt{Llama-2-7b}) and the instruction-tuned models (\ie \texttt{Llama-2-7b-chat}, \texttt{GPT-3.5-turbo}, and \texttt{GPT-4}).
The temperature is 0.
For the prompt optimizer, we use \texttt{GPT-3.5-turbo} and \texttt{GPT-4}.
The temperature is 1.0.
Unless otherwise specified, we use \texttt{Llama-2-7b-chat} as the task model and \texttt{GPT-3.5-turbo} as the prompt optimizer throughout experiments.
In the meta-prompt, we include 3 task examples and 7 historical task prompts.
For the initial task prompt, we use the original ones from \citet{Measuring-Hendrycks-2021-arXiv} for MMLU, ``Let's think step by step.'' from \citet{Large-Kojima-2022-NeurIPS} for GSM8K and BBH, and ``Let's solve the problem.'' from \citet{Large-Yang-2023-arXiv} for WSC and WebNLG.
At each step, the optimizer generates 8 candidates, and the best-performing one is selected.
The optimization lasts for at most 3 epochs with a batch size of 8.
We repeat all the experiments three times and report the average of the results.
The meta-prompts are detailed in Appendix~\ref{app:meta-prompt_exp}.

\section{B Extended Analysis}
\label{app:extended-analysis}

\begin{table*}[t]
\tiny
\centering
\resizebox{0.9 \textwidth}{!}{%
\begin{tabular}{c|c|cc}
\toprule
\multirow{2}{*}{\textbf{Method}}      & \multirow{2}{*}{\textbf{Prompt}}      & \multicolumn{2}{c}{\textbf{Accuracy}}   \\
                                      &                                        & MultiArith    & AQUA \\
\midrule
Zero-shot CoT                        & Let’s think step by step.             & 58.00  & 19.69    \\
\midrule
OPRO                                  & \begin{tabular}[c]{@{}c@{}} Take a deep breath and work on this problem step-by-step. \end{tabular}      & 59.33  & 20.47    \\
\midrule
GPO                                   & \begin{tabular}[c]{@{}c@{}} Thoroughly strategize and meticulously plan every single \\ aspect of your approach, leaving no room for oversight, \\  and give utmost attention to even the smallest details, \\ with the goal of achieving exceptional results and excelling. \end{tabular}      & \textbf{62.33} & \textbf{24.80}  \\             
\bottomrule
\end{tabular}%
}
\caption{Generalizability across datasets: Accuracies of optimal prompts for GSM8k on MultiArith and AQUA.}
\label{tab:generalizability}
\end{table*}

\subsection{B.1 The Convergence of LLM-based Prompt Optimization}
\label{app:convergence}

\begin{figure}[ht]
    \centering
    \begin{subfigure}[b]{0.49\linewidth}
        \centering
        \includegraphics[width=\textwidth]{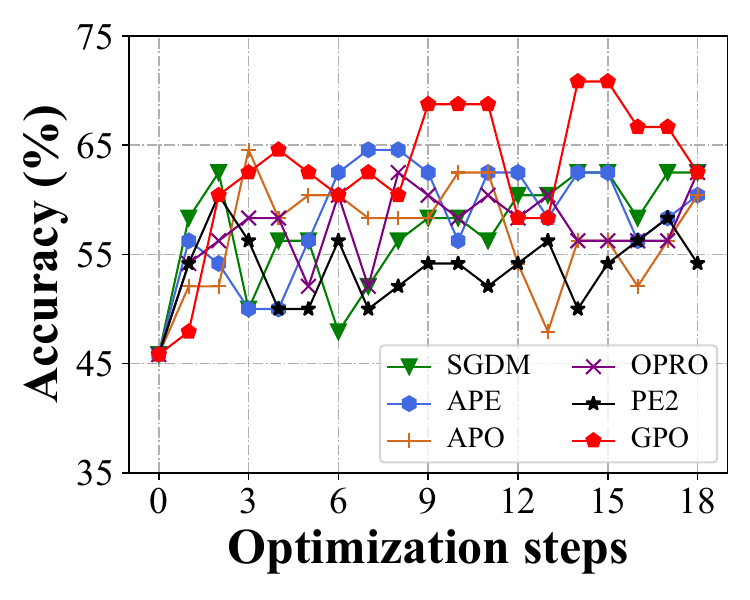}
        \caption{Optimization curve on Navigate dataset.}
        \label{fig:optim_n}
    \end{subfigure}
    \begin{subfigure}[b]{0.49\linewidth}
        \centering
        \includegraphics[width=\textwidth]{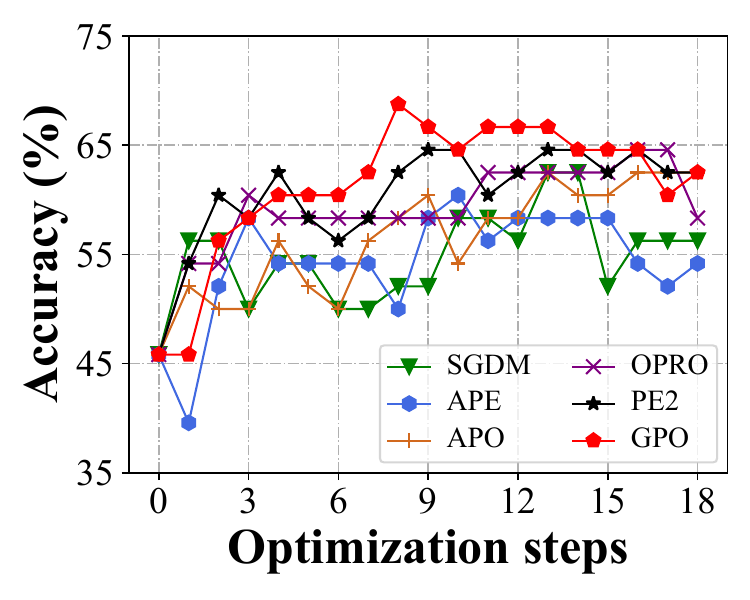}
        \caption{Optimization curve on Movie Recommendation dataset.}
        \label{fig:optim_mr}
    \end{subfigure}
    \centering
    \begin{subfigure}[b]{0.49\linewidth}
        \centering
        \includegraphics[width=\linewidth]{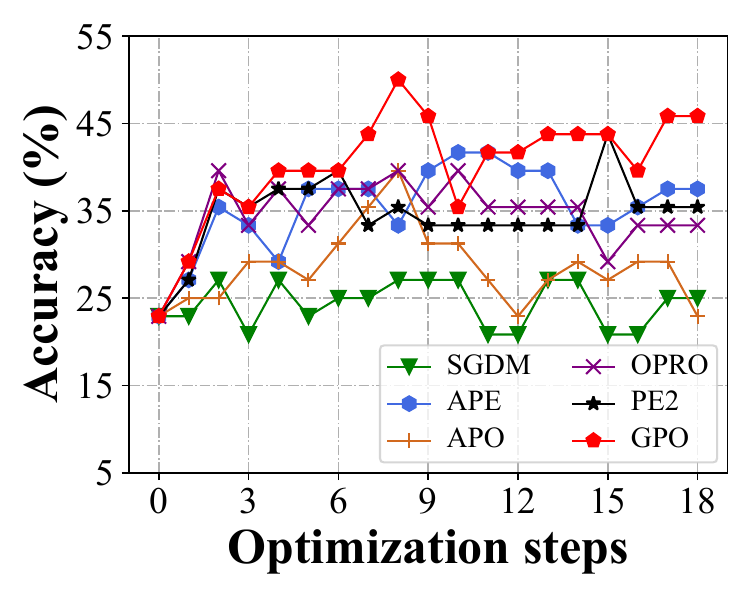}
        \caption{Optimization curve on Object Counting dataset.}
        \label{fig:optim_oc}
    \end{subfigure}
    \begin{subfigure}[b]{0.49\linewidth}
        \centering
        \includegraphics[width=\linewidth]{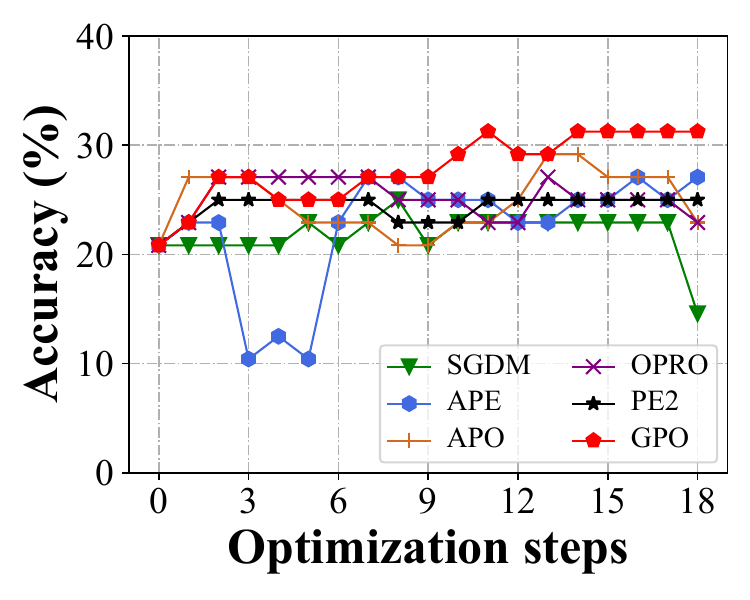}
        \caption{Optimization curve on Word Sorting dataset.}
        \label{fig:optim_ws}
    \end{subfigure}
\end{figure}

% \textcolor{blue}{
The convergence of LLM-based prompt optimization occurs when performance stabilizes at a plateau during the iterative process.
Here, we illustrate the entire optimization curve on the lite BBH benchmark in Figure~\ref{fig:optim_n}, Figure~\ref{fig:optim_mr}, Figure~\ref{fig:optim_oc}, and Figure~\ref{fig:optim_ws}.
It can be observed that all the prompt optimization methods tend to reach a stable state within 12 steps. 
Therefore, our experimental settings (\ie 3 epochs) are sufficient to demonstrate the effectiveness of GPO.
% Furthermore, \OURS demonstrates high efficiency in achieving rapid convergence and optimal performance.
% }

% \subsection{The Impact of Initial Prompts}
% \input{tables/initial_prompt}
% In our main experiment, we take ``Let's think step by step.'' as the initial prompt.
% In this part, we aim to explore the impact of initial task prompts.
% % \textcolor{blue}{
% Specifically, we consider prompts from four categories (instructive, misleading, irrelevant, and empty) following \citet{Large-Kojima-2022-NeurIPS}.
% % }
% % and select three prompts from each of the first three categories.}
% Table~\ref{tab:initial_prompt} presents the results on the lite BBH benchmark.
% In general, \OURS can consistently boost performance across various initial prompts.
% This observation underscores the versatility of \OURS as a prompt optimizer.
% Furthermore, the efficacy of optimization is more pronounced with relevant initial prompts (\ie instructive and misleading).
% It means that prompt initialization is still important, especially in conveying task-specific information.

\subsection{B.2 The Generalizability of Optimized Prompts}

% \textcolor{blue}{
To assess the generalizability of optimized prompts, following \citet{Large-Zhou-2023-ICLR}, we evaluate the prompts optimized on GSM8k~\cite{Training-Cobbe-2021-arXiv} by OPRO and \OURS on two additional math reasoning benchmarks MultiArith~\cite{2016-arXiv-MultiArith} and AQUA~\cite{2017-ACL-AQUA}.
Table~\ref{tab:generalizability} demonstrates that the optimized prompts by \OURS can be generalized well on these two benchmarks, which perform better than those by OPRO.
% }

\subsection{B.3 The Impact of Initial Prompts}
\begin{table}[t]
\centering
\small
\resizebox{\columnwidth}{!}{%
\begin{tabular}{c|c|c}
\toprule
\multicolumn{2}{c|}{\textbf{Initial prompt}}                                                            & \begin{tabular}[c]{@{}c@{}}\textbf{Accuracy}\\\textbf{(before / after)}\end{tabular} \\
\midrule
Default                         & \texttt{Let's think step by step.}                                         & 31.25 / 43.75  \\
\midrule
\multirow{4}{*}{Instructive}    & \texttt{Let’s think about this logically.}                                     & 32.64 / 44.40   \\
                                & \texttt{First,}                                                                & 33.85 / 43.15  \\
                                & \begin{tabular}[c]{@{}c@{}}\texttt{Let’s solve this problem}\\ \texttt{by splitting it into steps.}\end{tabular}                  & 30.38 / 37.85  \\
\midrule
\multirow{5}{*}{Misleading}     & \texttt{Don’t think. Just feel.}                                               & 29.34 / 39.59  \\
                                & \begin{tabular}[c]{@{}c@{}}\texttt{Let’s think step by step}\\ \texttt{but reach an incorrect answer.}\end{tabular}               & 26.22 / 41.15  \\
                                & \begin{tabular}[c]{@{}c@{}}\texttt{Let’s count the number of "a"}\\ \texttt{in the question}\end{tabular}                         & 26.73 / 36.11  \\
\midrule
\multirow{3}{*}{Irrelevant}     & \begin{tabular}[c]{@{}c@{}}\texttt{By the way, I found}\\ \texttt{a good restaurant nearby.}\end{tabular}                         & 30.03 / 34.89  \\
                                & \texttt{Abrakadabra!}                                                          & 29.69 / 35.42  \\
                                & \texttt{It’s a beautiful day.}                                                 & 30.55 / 38.02  \\
\midrule
Empty                           & Null                                                                  & 32.46 / 39.02 \\
% \midrule
% \multicolumn{2}{c|}{w/o optimization (Let's think step by step.)}                                       &       \\                      
\bottomrule
\end{tabular}%
}
\caption{The performance before/after prompt optimization with different initial prompts.}
\label{tab:initial_prompt}
\end{table}
In our main experiment, we take ``Let's think step by step.'' as the initial prompt.
In this part, we aim to explore the impact of initial task prompts.
% \textcolor{blue}{
Specifically, we consider prompts from four categories (instructive, misleading, irrelevant, and empty) following \citet{Large-Kojima-2022-NeurIPS}.
Table~\ref{tab:initial_prompt} presents the results on the lite BBH benchmark.
In general, \OURS can boost performance across various initial prompts.
This observation underscores the versatility of \OURS as a prompt optimizer.
Furthermore, the efficacy of optimization is more pronounced with relevant initial prompts (\ie instructive and misleading).
It means that prompt initialization is still important, especially in conveying task-specific information.

\subsection{B.4 Hyper-Parameter Analysis}
\begin{figure}[t]
    \centering
    \begin{subfigure}[b]{0.49\linewidth}
        \centering
        \includegraphics[width=\textwidth]{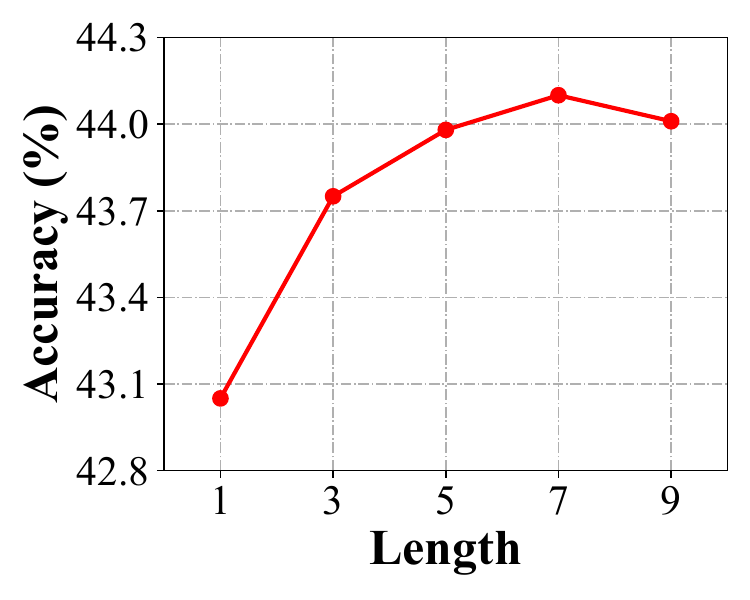}
        \caption{Trajectory length}
    \end{subfigure}
    \begin{subfigure}[b]{0.49\linewidth}
        \centering
        \includegraphics[width=\textwidth]{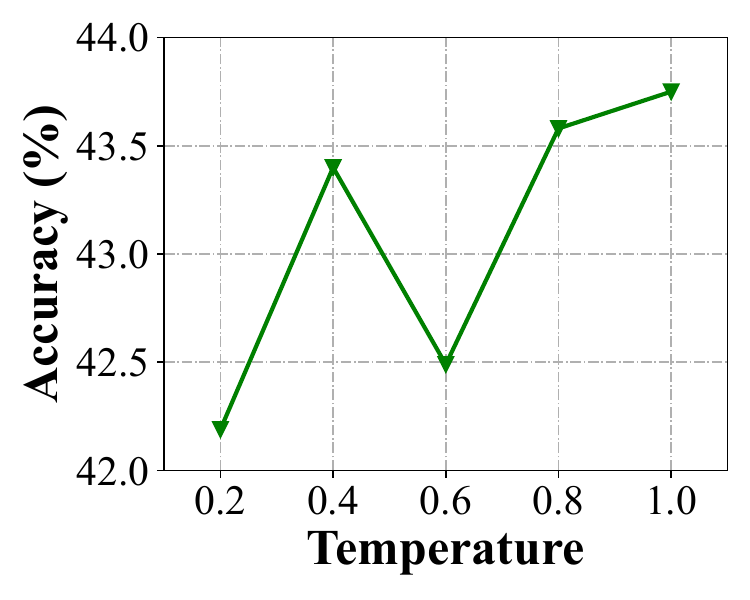}
        \caption{Temperature in \OURS}
    \end{subfigure}
    \caption{Performance comparison w.r.t. the temperature of the LLM in \OURS and the length of the trajectory.}
\label{fig:hyper-parameter}
\end{figure}

\OURS includes a few hyper-parameters to tune.
Here, we report the tuning results of two hyper-parameters on the lite BBH benchmark: the temperature of the prompt optimizer and the length of the trajectory.
The performance curve depicting these results is illustrated in Figure~\ref{fig:hyper-parameter}.
We can see that \OURS achieves the best performance when setting the length of the trajectory to 7.
A trajectory that is too short does not provide sufficient context for the LLM to engage in effective in-context learning. 
Conversely, a long trajectory may introduce excessive noise, negatively impacting performance.
In addition, the performance shows an upward trend as temperature increases, reaching its peak at 1.0.
It suggests the significance of exploration in optimizing task prompts.

\begin{table*}[t]
\centering
\tiny
\resizebox{0.9 \textwidth}{!}{%
\begin{tabular}{c|c}
\toprule
\textbf{Step}      & \textbf{Prompt}   \\
\midrule
0                  & \begin{tabular}[c]{@{}c@{}} Let’s think step by step. \end{tabular}  \\
\midrule
1                  & \begin{tabular}[c]{@{}c@{}} Plan your steps carefully. \end{tabular} \\
\midrule
2                  & \begin{tabular}[c]{@{}c@{}} Strategize your approach meticulously. \end{tabular} \\
\midrule
3                  & \begin{tabular}[c]{@{}c@{}} Strategically plan your approach down to the smallest details. \end{tabular} \\
\midrule
4                  & \begin{tabular}[c]{@{}c@{}} Carefully strategize and meticulously plan your approach down to the tiniest details. \end{tabular} \\
\midrule
5                  & \begin{tabular}[c]{@{}c@{}} Thoroughly strategize and meticulously plan every aspect of your approach, \\ ensuring attention to even the most minuscule details. \end{tabular} \\
\midrule
6                  & \begin{tabular}[c]{@{}c@{}} Strive to meticulously strategize and thoroughly plan each and every aspect of your approach, \\ leaving no room for oversight, and paying utmost attention to even the most infinitesimal details. \\ \end{tabular} \\
\midrule
9                  & \begin{tabular}[c]{@{}c@{}} Thoroughly strategize and meticulously plan every single aspect of your approach, ensuring there is no room \\ for oversight, and paying utmost attention to even the minutest details, in order to achieve perfect results. \end{tabular} \\
\midrule
12                 & \begin{tabular}[c]{@{}c@{}} Thoroughly strategize and meticulously plan every single aspect of your approach, leaving no room for oversight, \\ and give utmost attention to even the smallest details, with the goal of achieving exceptional results and excelling. \end{tabular} \\
\bottomrule
\end{tabular}%
}
\caption{Intermediate prompts optimized by \OURS on GSM8k dataset.}
\label{tab:interpretability1}
\end{table*}
\begin{table*}[t]
\centering
\tiny
\resizebox{0.9 \textwidth}{!}{%
\begin{tabular}{c|c}
\toprule
\textbf{Step}      & \textbf{Prompt}   \\
\midrule
0                  & \begin{tabular}[c]{@{}c@{}} Let’s think step by step. \end{tabular}  \\
\midrule
1                  & \begin{tabular}[c]{@{}c@{}} Thinking step by step will help us find the solution. \end{tabular} \\
\midrule
2                  & \begin{tabular}[c]{@{}c@{}} Sort the given list of words in alphabetical order. \end{tabular} \\
\midrule
3                  & \begin{tabular}[c]{@{}c@{}} Arrange the given list of words in ascending order based on their alphabetical order. \end{tabular} \\
\midrule
4                  & \begin{tabular}[c]{@{}c@{}} Arrange the given list of words in alphabetical order, from A to Z. \end{tabular} \\
\midrule
5                  & \begin{tabular}[c]{@{}c@{}} Sort the given words in alphabetical order, from A to Z. \end{tabular} \\
\midrule
6                  & \begin{tabular}[c]{@{}c@{}} Arrange the given words in alphabetical order, from A to Z. \end{tabular} \\
\midrule
9                  & \begin{tabular}[c]{@{}c@{}} Arrange the given list of words in ascending alphabetical order, starting from the word with the \\ lowest letter and ending with the word with the highest letter, disregarding case sensitivity. \end{tabular} \\
\midrule
12                 & \begin{tabular}[c]{@{}c@{}} Sort the given list of words in ascending alphabetical order, starting from the word with the lowest letter \\ and ending with the word with the highest letter, while considering both uppercase and lowercase letters. \end{tabular} \\
\bottomrule
\end{tabular}%
}
\caption{Intermediate prompts optimized by \OURS on Word Sorting dataset.}
\label{tab:interpretability2}
\end{table*}

\subsection{B.5 The Execution Time of LLM-based Prompt Optimizers}
\label{app:execution-time}

\begin{figure}[t]
    \centering
    \includegraphics[width=\linewidth]{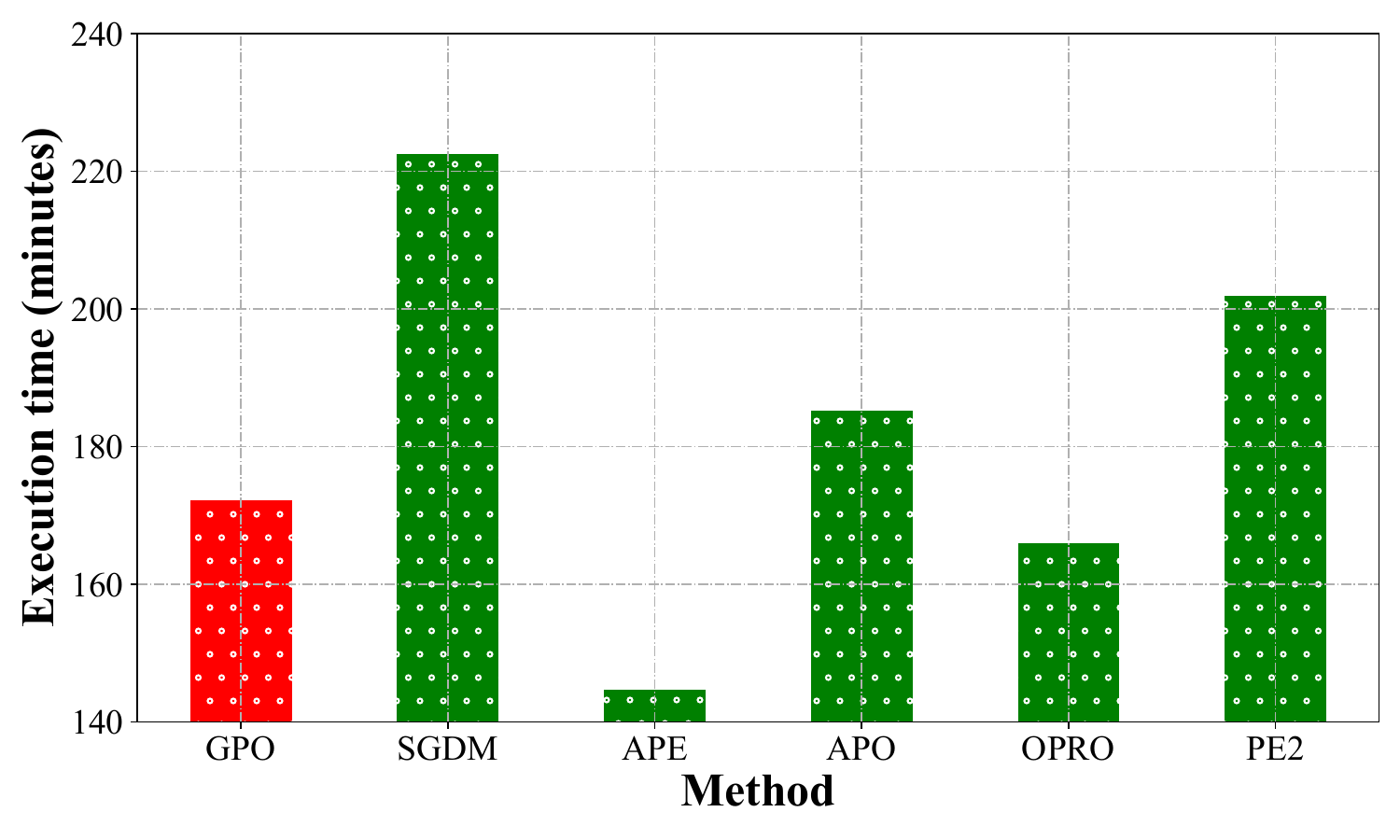}
    \caption{Execution time of LLM-based prompt optimizers}
\label{fig:execution-time}
\end{figure}

% \textcolor{blue}{
To further assess the computation costs of LLM-based prompt optimizers, we conduct five repetition experiments and record their average execution time.
Figure~\ref{fig:execution-time} shows the results on the lite BBH benchmark.
We can observe that the execution time is almost consistent with the trend of token consumption.
The average execution time of \OURS is much lower than SGDM, APO, and PE2, and comparable to OPRO.
% }

\section{C The Interpretability of Optimized Prompts}
\label{app:interpretability}

% \textcolor{blue}{
% In the above, we have observed that \OURS demonstrates its capability to effectively optimize prompts and achieve impressive performance.
To further enhance the interpretability of optimized prompts, we present the intermediate prompts optimized by GPO on GSM8k and Word Sorting dataset in Table~\ref{tab:interpretability1} and Table~\ref{tab:interpretability2}.
It can be observed that as the number of steps increases, the optimized prompts contain more task-related information 
(\eg ``Sort the given list of words in ascending alphabetical order, starting from the word with the lowest letter and ending with the word with the highest letter'') 
and fine-grained hints (\eg ``Thoroughly strategize and meticulously plan every single aspect of your approach, leaving no room for oversight, and give utmost attention to even the smallest detail'') 
to better solve the task.
In addition, we find that the optimized prompts do not contain information on specific task samples, leading to better generalizability and robustness.
% }

% ==================Navigate==================
\begin{table*}[t]
\centering
\small
\resizebox{\linewidth}{!}{%
\begin{tabular}{c|p{\linewidth}}
\toprule
\textbf{Methods}    & \multicolumn{1}{c}{\textbf{Optimized prompt}} \\
\midrule
SGDM                & Based on your current facing direction and any changes in direction, will following these step-by-step instructions, with explicit reference to direction and orientation, lead you back to the starting point? \\
\midrule
APE                 & Will following the given set of instructions result in returning to the starting point? \\
\midrule
APO                 & Based on the provided instructions, including the starting direction and the changes in direction throughout the sequence, determine if the person will return to the starting point. \\
\midrule
OPRO                & Think systematically and consider each step to determine the correct answer. \\
\midrule
PE2                 & Carefully analyze the given instructions step by step and determine if you will return to the starting point. \\
\midrule
\OURS               & Analyze the given step-by-step instructions in detail and determine if they will guide you back to the starting point. Carefully evaluate each instruction and vividly imagine the movements to make your decision. \\

\bottomrule
\end{tabular}%
}
\caption{Prompts optimized by different methods on the Navigate task.}
\label{tab:prompt_optimized_navigate}
\end{table*}
% ==================Movie Recommendation==================

\begin{table*}[t]
\centering
\small
\resizebox{\linewidth}{!}{%
\begin{tabular}{c|p{\linewidth}}
\toprule
\textbf{Methods}    & \multicolumn{1}{c}{\textbf{Optimized prompt}} \\
\midrule
SGDM                & What specific aspects of storytelling style, narrative structure, plot structure, character development, and narrative techniques should be considered when finding a movie similar to films like Pulp Fiction, Forrest Gump, Dances with Wolves, and The Usual Suspects? Analyze these elements in each movie's construction and compare them against the given list to determine similarity. Focus on shared storytelling elements rather than thematic or genre similarities. Keep in mind that the most important factor in determining similarity might not be the overall theme or genre. Provide a single letter as your answer. \\
\midrule
APE                 & Which of the following movies is most similar to the given movies based on their genre, themes, and plot elements? Choose the best option from the following choices: \\
\midrule
APO                 & Which movie from the following options is most similar to the given movies, taking into account specific themes, plot, genre, characters, setting, tone, and style? Please provide a ranked list of the relevant elements mentioned above in determining similarity between movies. Consider the highest ranked element as the primary criteria and subsequent elements as secondary criteria in determining similarity. Additionally, consider the overall popularity and critical acclaim of the movies when making your selection. This will help ensure more accurate predictions. \\
\midrule
OPRO                & Which of the given movies is most similar to the listed options? Consider the movies Batman, The Usual Suspects, The Silence of the Lambs, and Jurassic Park. Choose the option that closely matches the given movies. \\
\midrule
PE2                 & Select the movie option that is most closely related to the given list of movies after conducting a meticulous analysis. \\
\midrule
\OURS               & Thoroughly analyze the themes, genres, and narrative elements of each film to identify the movie that best aligns with the provided options. Make a well-informed decision based on your comprehensive evaluation. \\

\bottomrule
\end{tabular}%
}
\caption{Prompts optimized by different methods on the Movie Recommendation task.}
\label{tab:prompt_optimized_movie_recommendation}
\end{table*}
% ==================Object counting==================
\begin{table*}[t]
\centering
\small
\resizebox{\linewidth}{!}{%
\begin{tabular}{c|p{\linewidth}}
\toprule
\textbf{Methods}    & \multicolumn{1}{c}{\textbf{Optimized prompt}} \\
\midrule
SGDM                & Count the number of items \\
\midrule
APE                 & Develop a systematic approach to accurately determine the total number of items by counting each individual item separately and recording their corresponding quantities. \\
\midrule
APO                 & What is the total count of mentioned items, considering each item individually without categorization and counting duplicates as separate items? Please provide the correct count as your output. \\
\midrule
OPRO                & How can we solve the problem by breaking it down step by step? \\
\midrule
PE2                 & By employing a systematic and thorough approach, meticulously analyze each item in a step-by-step manner to precisely determine the total count. \\
\midrule
\OURS               & Let's break down the problem systematically by deconstructing it into individual steps and accurately computing the total number of objects. \\
\bottomrule
\end{tabular}%
}
\caption{Prompts optimized by different methods on the Object Counting task.}
\label{tab:prompt_optimized_object_counting}
\end{table*}
% ==================Word Sorting==================
\begin{table*}[t]
\centering
\small
\resizebox{\linewidth}{!}{%
\begin{tabular}{c|p{\linewidth}}
\toprule
\textbf{Methods}    & \multicolumn{1}{c}{\textbf{Optimized prompt}} \\
\midrule
SGDM                & Sort the given words, alphabetically, in case-sensitive order, considering the entire word, including all characters. The sorting should be done in ascending order based on the lowercase versions of the words, while preserving the original case and considering the entire word. \\
\midrule
APE                 & How would you sort the given list of words in alphabetical order, considering both uppercase and lowercase letters? \\
\midrule
APO                 & Sort the given list of words in lowercase alphabetical order, taking into account all characters including special characters and numbers. Ensure that the sorting process is case-insensitive and considers all characters, including special characters and numbers. Convert all letters to lowercase before sorting. Consider special characters and numbers in the sorting process. \\
\midrule
OPRO                & Arrange the given words in alphabetical order by considering only the first letter of each word and exclude punctuation or special characters. \\
\midrule
PE2                 & Analyze the given words and provide a sorted list in alphabetical order. \\
\midrule
\OURS               & Arrange the following words in alphabetical order: \\

\bottomrule
\end{tabular}%
}
\caption{Prompts optimized by different methods on the Word Sorting task.}
\label{tab:prompt_optimized_word_sorting}
\end{table*}

\section{D Prompts Optimized by Different Methods}

In this part, we present the prompts optimized by all the methods on the lite BBH benchmark in Table~\ref{tab:prompt_optimized_navigate}, Table~\ref{tab:prompt_optimized_movie_recommendation}, Table~\ref{tab:prompt_optimized_object_counting} and Table~\ref{tab:prompt_optimized_word_sorting}.

\section{E Detailed Experiment Results}
\begin{table*}[t]
\centering
\tiny
\resizebox{0.95 \textwidth}{!}{
    \begin{tabular}{l|cc|ccccc|c}
    \toprule
\textbf{MMLU Tasks} & \textbf{Empty} & \textbf{CoT} & \textbf{SGDM} & \textbf{APE} & \textbf{APO} & \textbf{OPRO} & \textbf{PE2}  &  \textbf{\OURS} \\
\midrule
abstract algebra               & 27.12 & 22.03  & 27.33 & 32.20 & 27.12 & 25.42 & 30.51 & 32.20 \\
anatomy                        & 28.33 & 26.67  & 23.33 & 30.00 & 28.33 & 31.67 & 30.00 & 36.67 \\
astronomy                      & 35.00 & 36.67  & 38.33 & 33.33 & 35.00 & 43.33 & 41.67 & 41.67 \\
business ethics                & 37.29 & 35.59  & 31.67 & 27.12 & 37.29 & 35.59 & 45.76 & 37.29 \\
clinical knowledge             & 36.67 & 41.67  & 43.33 & 38.33 & 40.00 & 38.33 & 36.67 & 38.33 \\
college biology                & 36.67 & 38.33  & 45.00 & 43.33 & 43.33 & 45.00 & 45.00 & 48.33 \\
college chemistry              & 37.29 & 33.90  & 33.90 & 35.59 & 35.59 & 40.68 & 37.29 & 38.98 \\
college computer science       & 37.29 & 38.98  & 33.33 & 40.68 & 35.59 & 38.98 & 38.98 & 37.29 \\
college mathematics            & 30.51 & 27.12  & 38.98 & 30.51 & 38.98 & 32.20 & 32.20 & 33.90 \\
college medicine               & 30.00 & 40.00  & 31.45 & 33.33 & 31.67 & 28.33 & 26.67 & 31.67 \\
college physics                & 23.33 & 31.67  & 36.67 & 36.67 & 35.00 & 33.90 & 38.98 & 31.67 \\
computer security              & 44.07 & 37.29  & 44.07 & 38.98 & 42.37 & 38.98 & 33.90 & 35.59 \\
conceptual physics             & 28.33 & 38.33  & 35.00 & 41.67 & 35.00 & 36.67 & 35.00 & 33.33 \\
econometrics                   & 25.00 & 28.33  & 23.33 & 28.33 & 26.67 & 31.67 & 23.33 & 26.67 \\
electrical engineering         & 35.00 & 31.67  & 35.00 & 35.00 & 31.67 & 40.00 & 30.00 & 40.00 \\
elementary mathematics         & 33.33 & 33.33  & 36.67 & 33.33 & 33.33 & 36.67 & 33.33 & 36.67 \\
formal logic                   & 40.00 & 46.67  & 35.00 & 36.67 & 38.33 & 45.00 & 43.33 & 36.67 \\
global facts                   & 40.68 & 32.20  & 35.33 & 35.59 & 38.98 & 35.59 & 23.73 & 35.59 \\
high school biology            & 28.33 & 43.33  & 30.00 & 33.33 & 26.67 & 31.67 & 33.33 & 31.67 \\
high school chemistry          & 28.33 & 26.67  & 28.67 & 31.67 & 31.67 & 23.33 & 30.00 & 35.00 \\
high school computer science   & 20.34 & 28.81  & 23.73 & 20.34 & 30.51 & 20.34 & 23.73 & 22.03 \\
high school european history   & 41.67 & 48.33  & 40.00 & 46.67 & 41.67 & 45.00 & 48.33 & 48.33 \\
high school geography          & 48.33 & 50.00  & 40.00 & 45.00 & 45.00 & 50.00 & 51.67 & 51.67 \\
high school government and politics & 45.00 & 45.00  & 35.00 & 43.33 & 45.00 & 41.67 & 50.00 & 50.00 \\
high school macroeconomics     & 45.00 & 30.00  & 46.67 & 36.67 & 46.67 & 43.33 & 46.67 & 48.33 \\
high school mathematics        & 18.33 & 21.67  & 18.33 & 25.00 & 18.33 & 25.00 & 20.00 & 23.33 \\
high school microeconomics     & 31.67 & 33.33  & 35.00 & 35.00 & 38.33 & 31.67 & 35.00 & 41.67 \\
high school physics            & 31.67 & 35.00  & 30.00 & 28.33 & 28.33 & 35.00 & 33.33 & 31.67 \\
high school psychology         & 48.33 & 53.33  & 50.00 & 55.00 & 58.33 & 53.33 & 55.00 & 53.33 \\
high school statistics         & 36.67 & 35.00  & 35.00 & 30.00 & 36.67 & 40.00 & 35.00 & 36.67 \\
high school us history         & 41.67 & 38.33  & 38.33 & 38.33 & 40.00 & 41.67 & 40.00 & 38.33 \\
high school world history      & 43.33 & 46.67  & 41.67 & 41.67 & 43.33 & 45.00 & 40.00 & 43.33 \\
human aging                    & 28.33 & 26.67  & 38.33 & 38.33 & 35.00 & 30.00 & 41.67 & 40.00 \\
human sexuality                & 30.00 & 33.33  & 31.67 & 38.33 & 40.00 & 38.33 & 45.00 & 40.00 \\
international law              & 38.33 & 40.00  & 40.00 & 41.33 & 38.33 & 33.33 & 40.00 & 38.33 \\
jurisprudence                  & 35.00 & 40.00  & 38.67 & 38.33 & 33.33 & 23.33 & 43.33 & 38.33 \\
logical fallacies              & 40.00 & 30.00  & 38.33 & 40.00 & 40.00 & 40.00 & 30.00 & 35.00 \\
machine learning               & 30.00 & 31.67  & 31.67 & 36.67 & 36.67 & 41.67 & 40.00 & 38.33 \\
management                     & 40.00 & 46.67 & 40.00 & 41.67 & 48.33 & 41.67 & 46.67 & 43.33 \\
marketing                      & 43.33 & 46.67 & 38.33 & 46.67 & 45.00 & 40.67 & 48.67 & 56.67 \\
medical genetics               & 40.68 & 47.46 & 47.55 & 45.76 & 40.68 & 38.98 & 42.37 & 35.59 \\
miscellaneous                  & 43.33 & 40.00 & 36.67 & 36.67 & 33.33 & 40.00 & 36.67 & 40.00 \\
moral disputes                 & 26.67 & 25.00 & 38.33 & 28.33 & 21.67 & 28.33 & 30.00 & 30.00 \\
moral scenarios                & 30.00 & 11.67 & 30.00 & 31.67 & 45.00 & 33.33 & 26.67 & 31.67 \\
nutrition                      & 45.00 & 36.67 & 40.00 & 48.33 & 45.00 & 43.33 & 48.33 & 41.67 \\
philosophy                     & 28.33 & 33.33 & 45.00 & 40.00 & 40.00 & 50.00 & 36.67 & 45.00 \\
prehistory                     & 25.00 & 28.33 & 23.33 & 31.67 & 26.67 & 28.33 & 20.00 & 28.33 \\
professional accounting        & 26.67 & 23.33 & 38.33 & 25.00 & 28.33 & 45.00 & 26.67 & 30.00 \\
professional law               & 33.33 & 33.33 & 33.33 & 35.00 & 31.67 & 35.00 & 31.67 & 33.33 \\
professional medicine          & 45.00 & 38.33 & 43.33 & 43.33 & 45.00 & 38.33 & 40.00 & 43.33 \\
professional psychology        & 38.33 & 35.00 & 48.33 & 33.33 & 35.00 & 40.00 & 35.00 & 43.33 \\
public relations               & 35.00 & 36.67 & 38.33 & 41.67 & 41.67 & 36.78 & 38.33 & 40.00 \\
security studies               & 48.33 & 46.67 & 46.67 & 45.00 & 56.67 & 56.67 & 48.33 & 50.00 \\
sociology                      & 46.67 & 45.00 & 51.67 & 46.67 & 48.33 & 50.00 & 51.67 & 55.00 \\
us foreign policy              & 59.32 & 55.93 & 55.93 & 55.93 & 57.63 & 49.15 & 57.63 & 59.32 \\
virology                       & 26.67 & 25.00 & 30.00 & 33.33 & 28.33 & 26.67 & 33.33 & 28.33 \\
world religions                & 51.67 & 60.00 & 56.67 & 53.33 & 43.33 & 55.00 & 63.33 & 58.33 \\
\midrule
Average                        & 35.96 & 36.36 & 37.20 & 37.50 & 37.71 & 38.05 & 38.07 & \textbf{39.14} \\
    \bottomrule
    \end{tabular}
    }
\caption{\label{tab: mmlu-detail} Detailed Results (Accuracy\%) on MMLU.}
\end{table*}

\begin{table*}[t]
\centering
\tiny
\resizebox{\textwidth}{!}{
\begin{tabular}{l|cc|ccccc|c}
\toprule
\textbf{BBH Tasks} & \textbf{Empty} & \textbf{CoT} & \textbf{SGDM} & \textbf{APE} & \textbf{APO} & \textbf{OPRO} & \textbf{PE2}  &  \textbf{\OURS}\\
\hline
Boolean Expressions & 38.89 & 43.06 & 48.61 & 58.33 & 43.75 & 42.36 & 45.14 & 51.39 \\
Causal Judgement & 28.04 & 23.36 & 46.73 & 44.86 & 42.99 & 49.53 & 50.47 & 47.66 \\
Date Understanding & 29.86 & 36.11 & 34.03 & 22.22 & 35.42 & 34.72 & 34.72 & 33.33 \\
Disambiguation QA & 53.47 & 37.50 & 46.53 & 33.33 & 40.97 & 29.17 & 34.03 & 47.92 \\
Dyck Languages & 1.39 & 2.78 & 5.56 & 7.64 & 2.78 & 4.86 & 2.08 & 4.17 \\
Formal Fallacies & 59.03 & 54.17 & 55.56 & 54.86 & 54.86 & 48.61 & 45.83 & 55.56 \\
Geometric Shapes & 6.94 & 14.58 & 24.31 & 22.92 & 29.17 & 22.22 & 27.08 & 28.47 \\
Hyperbaton & 49.31 & 63.89 & 63.19 & 61.81 & 61.55 & 75.00 & 65.97 & 62.50 \\
Logical Deduction (five objects) & 20.83 & 22.92 & 25.69 & 22.92 & 24.31 & 23.91 & 25.00 & 23.61 \\
Logical Deduction (seven objects) & 19.44 & 17.36 & 13.89 & 12.50 & 13.19 & 18.75 & 16.67 & 17.36 \\
Logical Deduction (three objects) & 45.14 & 36.11 & 37.50 & 42.36 & 34.72 & 36.81 & 36.81 & 41.67 \\
Movie Recommendation & 27.08 & 27.78 & 48.61 & 50.69 & 48.61 & 52.08 & 54.17 & 50.69 \\
Multi-Step Arithmetic & 0.00 & 2.08 & 0.69 & 2.78 & 2.08 & 0.00 & 3.47 & 2.78 \\
Navigate & 44.44 & 43.75 & 59.03 & 51.39 & 49.57 & 49.84 & 47.22 & 51.93 \\
Object Counting & 32.64 & 36.81 & 36.81 & 43.75 & 50.69 & 44.44 & 45.83 & 50.86 \\
Penguins in a Table & 31.71 & 19.51 & 29.27 & 26.83 & 20.73 & 24.39 & 20.73 & 26.83 \\
Reasoning about Colored Objects & 29.17 & 22.92 & 22.92 & 21.53 & 21.53 & 25.00 & 21.53 & 20.83 \\
Ruin Names & 27.78 & 30.56 & 23.61 & 30.56 & 29.17 & 27.78 & 29.86 & 31.94 \\
Salient Translation Error Detection & 15.97 & 13.19 & 7.64 & 15.28 & 18.75 & 19.44 & 23.61 & 18.75 \\
Snarks & 46.08 & 59.80 & 62.75 & 64.71 & 52.94 & 56.86 & 59.80 & 63.73 \\
Sports Understanding & 54.86 & 59.72 & 59.72 & 52.08 & 63.19 & 59.03 & 63.19 & 62.50 \\
Temporal Sequences & 20.83 & 22.92 & 27.78 & 19.44 & 29.17 & 22.22 & 25.00 & 23.61 \\
Tracking Shuffled Objects (five objs) & 18.06 & 10.42 & 11.81 & 12.84 & 10.42 & 16.67 & 12.50 & 17.36 \\
Tracking Shuffled Objects (seven objs) & 14.58 & 9.72 & 13.19 & 17.36 & 10.42 & 17.36 & 13.19 & 16.67 \\
Tracking Shuffled Objects (three objs) & 36.11 & 32.64 & 30.56 & 29.86 & 34.72 & 29.86 & 33.33 & 34.03 \\
Web of Lies & 47.22 & 47.22 & 45.14 & 44.44 & 52.08 & 45.83 & 44.44 & 49.01 \\
Word Sorting & 25.00 & 16.67 & 18.08 & 22.22 & 12.50 & 22.19 & 20.83 & 21.53 \\
\hline
All Tasks (avg) & 30.51 & 29.91 & 33.30 & 32.94 & 32.97 & 33.29 & 33.43 & \textbf{35.43} \\
\bottomrule
\end{tabular}
}
\caption{\label{tab: bbh-detail}Detailed Results (Accuracy (\%)) on the BBH benchmark.}
\end{table*}

In this section, we provide the detailed results of the experiment on MMLU in Table~\ref{tab: mmlu-detail} and BBH in Table~\ref{tab: bbh-detail}.

\section{F Meta-Prompt}
\label{app:meta-prompt}

\subsection{F.1 Analogical Analysis}
\label{app:meta-prompt_analysis}
Here are the meta-prompts we used in the Analogical Analysis Section of the main body.

\begin{center}
\begin{tcolorbox}[colback=blue!5!white,colframe=blue!55!black,width=0.46\textwidth,title={Prompt+performance.}]
    {
    \small
    Below is the current prompt with its score. The score ranges from 0 to 100, and higher score indicates better quality.\\
    Prompt: \{current prompt\}\\
    Score: \{current prompt score\}
    }
    % \vspace{-5pt}
\end{tcolorbox}
\end{center}

\begin{center}
\begin{tcolorbox}[colback=blue!5!white,colframe=blue!55!black,width=0.46\textwidth,title={Prompt+performance+reflection}]
    {
    \small
    Your task is to point out the problems with the current prompt based on the wrong examples.\\
    \\
    The current prompt is:\\
    \{current prompt\}\\
    \\
    But this prompt gets the following examples wrong.\\
    You should analyze the differences between wrong predictions and ground truth answers, and carefully consider why this prompt led to incorrect predictions.\\
    \\
    Below are the task examples with Queston, Wrong prediction, and Ground truth answer.\\
    \\
    \{error demonstrations\}\\
    \\
    Give a reason why the prompt could have gotten these examples wrong.\\
    Wrap the reason with START and END.
    }
    % \vspace{-5pt}
\end{tcolorbox}
\end{center}

\begin{center}
\begin{tcolorbox}[colback=blue!5!white,colframe=blue!55!black,width=0.46\textwidth,title={Summarization-based trajectory}]
    {
    \small
    Your task is to integrate the problems in the previous prompt and the current prompt.\\
    \\
    Below are the problems that arose from the previous prompts.\\
    \{previous problems\}\\
    \\
    Below are the problems of the current prompt.\\
    \{current problem\}\\
    \\
    You should integrate the problems of the previous prompt and the current prompt.\\
    Wrap the integrated problems with START and END.
    }
\end{tcolorbox}
\end{center}

\begin{center}
\begin{tcolorbox}[colback=blue!5!white,colframe=blue!55!black,width=0.46\textwidth,title={Retrieval-based trajectory}]
    {
    \small
    Below are the previous prompts with their scores. The score ranges from 0 to 100, and higher scores indicate better quality. \\
    \\
    Prompt: \{prompt1\} \\
    Score: \{score1\} \\
    \\
    Prompt: \{prompt2\} \\
    Score: \{score2\} \\
    \\
    Prompt: \{prompt3\} \\
    Score: \{score3\} \\
    \ldots
    }
\end{tcolorbox}
\end{center}

\begin{center}
\begin{tcolorbox}[colback=blue!5!white,colframe=blue!55!black,width=0.46\textwidth,title={Editing-based refinement}]
    {
    \small
    Your task is to modify the current prompt to replace Prompt \\
    \\
    Below is the current prompt with its score. The score ranges from 0 to 100, and higher score indicates better quality.\\
    Prompt: \{current prompt\}\\
    Score: \{current prompt score\}\\
    \\
    The current prompt is:\\
    \{current prompt\}\\
    \\
    The following exemplars show how to apply the prompt: you replace Prompt in each input with your new prompt, then read the input and give an output. We say your output is wrong if it is different from the given output, and we say your output is correct if they are the same.\\
    \\
    \{task examples\}\\
    \\
    Modify the current prompt and get a new improved prompt to replace Prompt \{prompt position description\} in the task examples.\\
    Wrap the modified prompt with START and END.
    }
\end{tcolorbox}
\end{center}

\begin{center}
\begin{tcolorbox}[colback=blue!5!white,colframe=blue!55!black,width=0.46\textwidth,title={Generation-based refinement}]
    {
    \small
    Your task is to write a prompt to replace Prompt.\\
    \\
    Below is the current prompt with its score. The score ranges from 0 to 100, and higher score indicates better quality.\\
    Prompt: \{current prompt\}\\
    Score: \{current prompt score\}\\
    \\
    The current prompt is:\\
    \{current prompt\}\\
    \\
    The following exemplars show how to apply the prompt: you replace Prompt in each input with your new prompt, then read the input and give an output. We say your output is wrong if it is different from the given output, and we say your output is correct if they are the same.\\
    \\
    \{task examples\}\\
    \\
    Write a new improved prompt to replace Prompt \{prompt position description\} in the task examples.\\
    Wrap the new prompt with START and END.
    }
\end{tcolorbox}
\end{center}

\newpage
\subsection{F.2 Experiment}
\label{app:meta-prompt_exp}
Here are the meta-prompts we used in the Experiments Section of the main body.

\begin{center}
\begin{tcolorbox}[colback=blue!5!white,colframe=blue!55!black,width=0.46\textwidth,title={\textit{APE}}]
    {
    \large \textbf{The meta-prompt for update:} \\
    \\
    \small
    Your task is to write a prompt to replace Prompt.\\
    \\
    Below is the current prompt with its score. The score ranges from 0 to 100, and higher score indicates better quality.\\
    Prompt: \{current prompt\}\\
    Score: \{current prompt score\}\\
    \\
    The current prompt is:\\
    \{current prompt\}\\
    \\
    The following exemplars show how to apply the prompt: you replace Prompt in each input with your new prompt, then read the input and give an output. We say your output is wrong if it is different from the given output, and we say your output is correct if they are the same.\\
    \\
    \{task examples\}\\
    \\
    Write a new improved prompt to replace Prompt \{prompt position description\} in the task examples.\\
    Wrap the new prompt with START and END.
    }
\end{tcolorbox}
\end{center}

\newpage

\begin{center}
\begin{tcolorbox}[colback=blue!5!white,colframe=blue!55!black,width=0.46\textwidth,title={\textit{APO}}]
    {
    \large \textbf{The meta-prompt for gradient:} \\
    \\
    \small
    Your task is to point out the problems with the current prompt based on the wrong examples.\\
    \\
    The current prompt is:\\
    \{current prompt\}\\
    \\
    But this prompt gets the following examples wrong.\\
    You should analyze the differences between wrong predictions and ground truth answers, and carefully consider why this prompt led to incorrect predictions.\\
    \\
    Below are the task examples with Queston, Wrong prediction, and Ground truth answer.\\
    \\
    \{error demonstrations\}\\
    \\
    Give a reason why the prompt could have gotten these examples wrong.\\
    Wrap the reason with START and END.
    \\
    \\
    \large \textbf{The meta-prompt for update:} \\
    \\
    \small
    Your task is to modify the current prompt to replace Prompt.\\
    \\
    Below is the current prompt with its score. The score ranges from 0 to 100, and higher score indicates better quality.\\
    Prompt: \{current prompt\}\\
    Score: \{current prompt score\}\\
    \\
    The current prompt is:\\
    \{current prompt\}\\
    \\
    Below are the problems with this prompt.\\
    \{problems\}\\
    \\
    The following exemplars show how to apply the prompt: you replace Prompt in each input with your new prompt, then read the input and give an output. We say your output is wrong if it is different from the given output, and we say your output is correct if they are the same.\\
    \\
    \{task examples\}\\
    \\
    Modify the current prompt and get a new improved prompt to replace Prompt \{prompt position description\} in the task examples.\\
    Wrap the modified prompt with START and END.
    }
\end{tcolorbox}
\end{center}

\begin{center}
\begin{tcolorbox}[colback=blue!5!white,colframe=blue!55!black,width=0.46\textwidth,title={\textit{OPRO}}]
    {
    \large \textbf{The meta-prompt for update:} \\
    \\
    \small
    Your task is to write a prompt to replace Prompt.\\
    \\
    Below are the previous prompts with their scores. The score ranges from 0 to 100, and higher scores indicate better quality.\\
    \\Prompt: \{prompt1\} \\ Score: \{score1\} \\
    \\Prompt: \{prompt2\} \\ Score: \{score2\} \\
    \\Prompt: \{prompt3\} \\ Score: \{score3\} \\
    \ldots\\
    \\
    The current prompt is:\\
    \{current prompt\}\\
    \\
    The following exemplars show how to apply the prompt: you replace Prompt in each input with your new prompt, then read the input and give an output. We say your output is wrong if it is different from the given output, and we say your output is correct if they are the same.\\
    \\
    \{task examples\}\\
    \\
    Carefully analyze the previous prompts and their scores, and write a new improved prompt to replace Prompt \{prompt position description\} in the task examples.\\
    Wrap the new prompt with START and END.
    }
\end{tcolorbox}
\end{center}

\begin{center}
\begin{tcolorbox}[colback=blue!5!white,colframe=blue!55!black,width=0.46\textwidth,title={\textit{PE2}}]
    {
    \large \textbf{The meta-prompt for gradient:} \\
    \\
    \small
    Your task is to point out the problems with the current prompt based on the wrong examples.\\
    \\
    The current prompt is:\\
    \{current prompt\}\\
    But this prompt gets the following examples wrong.\\
    You should analyze the differences between wrong predictions and ground truth answers, and carefully consider why this prompt led to incorrect predictions.\\
    \\
    Below are the task examples with Queston, Wrong prediction, and Ground truth answer.\\
    \{error demonstrations\}\\
    \\
    Give a reason why the prompt could have gotten these examples wrong.\\
    Wrap the reason with START and END.
    \\
    \\
    \large \textbf{The meta-prompt for update:} \\
    \\
    \small
    Your task is to write a prompt to replace Prompt.\\
    \\
    Below are the previous prompts with their scores. The score ranges from 0 to 100, and higher scores indicate better quality.\\
    \\Prompt: \{prompt1\} \\ Score: \{score1\} \\
    \\Prompt: \{prompt2\} \\ Score: \{score2\} \\
    \\Prompt: \{prompt3\} \\ Score: \{score3\} \\
    \ldots\\
    \\
    The current prompt is:\\
    \{current prompt\}\\
    \\
    Below are the problems with this prompt.\\
    \{problems\}\\
    \\
    The following exemplars show how to apply the prompt: you replace Prompt in each input with your new prompt, then read the input and give an output. We say your output is wrong if it is different from the given output, and we say your output is correct if they are the same.\\
    \\
    \{task examples\}\\
    \\
    Carefully analyze the previous prompts and their scores, and write a new improved prompt to replace Prompt \{prompt position description\} in the task examples.\\
    You are allowed to change up to \{modified word number\} words in the current prompt.\\
    Wrap the new prompt with START and END.
    }
\end{tcolorbox}
\end{center}

\begin{center}
\begin{tcolorbox}[colback=blue!5!white,colframe=blue!55!black,width=0.46\textwidth,title=\textit{SGDM}]
    {
    \large \textbf{The meta-prompt for gradient:} \\
    \\
    \small
    Your task is to point out the problems with the current prompt based on the wrong examples.\\
    \\
    The current prompt is:\\
    \{current prompt\}\\
    But this prompt gets the following examples wrong.\\
    You should analyze the differences between wrong predictions and ground truth answers, and carefully consider why this prompt led to incorrect predictions.\\
    \\
    Below are the task examples with Queston, Wrong prediction, and Ground truth answer.\\
    \{error demonstrations\}\\
    \\
    Give a reason why the prompt could have gotten these examples wrong.\\
    Wrap the reason with START and END.\\
    \\
    \large \textbf{The meta-prompt for momentum:} \\
    \\
    \small
    Your task is to integrate the problems in the previous prompt and the current prompt.\\
    \\
    Below are the problems that arose from the previous prompts.\\
    \{previous problems\}\\
    \\
    Below are the problems of the current prompt.\\
    \{current problem\}\\
    You should integrate the problems of the previous prompt and the current prompt.\\
    Wrap the integrated problems with START and END. \\
    \\
    \large \textbf{The meta-prompt for update:} \\
    \\
    \small
    Your task is to modify the current prompt to replace Prompt.\\
    Below is the current prompt with its score. The score ranges from 0 to 100, and higher score indicates better quality.\\
    Prompt: \{current prompt\}\\
    Score: \{current prompt score\}\\
    The current prompt is:\\
    \{current prompt\}\\
    \\
    Below are the problems with this prompt.\\
    \{problems\}\\
    \\
    The following exemplars show how to apply the prompt: you replace Prompt in each input with your new prompt, then read the input and give an output. We say your output is wrong if it is different from the given output, and we say your output is correct if they are the same.\\
    \\
    \{task examples\}\\
    \\
    Modify the current prompt and get a new improved prompt to replace Prompt \{prompt position description\} in the task examples.\\
    Wrap the modified prompt with START and END.
    }
\end{tcolorbox}
\end{center}

\begin{center}
\begin{tcolorbox}[colback=blue!5!white,colframe=blue!55!black,width=0.46\textwidth,title={\textit{\OURS}}]
    {
    \large \textbf{The meta-prompt for update:} \\
    \\
    \small
    Your task is to write a prompt to replace Prompt.\\
    \\
    Below are the previous prompts with their scores. The score ranges from 0 to 100, and higher scores indicate better quality.\\
    \\Prompt: \{prompt1\} \\ Score: \{score1\} \\
    \\Prompt: \{prompt2\} \\ Score: \{score2\} \\
    \\Prompt: \{prompt3\} \\ Score: \{score3\} \\
    \ldots\\
    \\
    The current prompt is:\\
    \{current prompt\}\\
    \\
    The following exemplars show how to apply the prompt: you replace Prompt in each input with your new prompt, then read the input and give an output. We say your output is wrong if it is different from the given output, and we say your output is correct if they are the same.\\
    \\
    \{task examples\}\\
    \\
    Carefully analyze the previous prompts and their scores, and write a new improved prompt to replace Prompt \{prompt position description\} in the task examples.\\
    You are allowed to change up to \{modified word number\} words in the current prompt.\\
    Wrap the new prompt with START and END.
    }
\end{tcolorbox}
\end{center}

\end{document}